\documentclass[pdflatex,sn-mathphys-num]{sn-jnl}

\usepackage{graphicx}%
\usepackage{multirow}%
\usepackage{amsmath,amssymb,amsfonts}%
\usepackage{amsthm}%
\usepackage{mathrsfs}%
\usepackage[title]{appendix}%
\usepackage{xcolor}%
\usepackage{textcomp}%
\usepackage{manyfoot}%
\usepackage{booktabs}%
\usepackage{algorithm}%
\usepackage{algorithmicx}%
\usepackage{algpseudocode}%
\usepackage{listings}%

\newcommand*{\xml}[1]{\texttt{<#1>}}

\usepackage{graphicx}
\usepackage{booktabs}
\usepackage{multirow}
\usepackage{enumitem}
\usepackage{tabularx}
\usepackage{makecell}
\usepackage{float}
\usepackage{comment}
\usepackage{caption}
\usepackage{subcaption}
\usepackage{lscape}
\usepackage{xurl}
\usepackage{xcolor}
\usepackage{ragged2e}
\usepackage{array}

\theoremstyle{thmstyleone}%
\theoremstyle{thmstyletwo}%

\theoremstyle{thmstylethree}%

\begin{document}

\title[Article Title]{Prompting Encoder Models for Zero-Shot Classification: A Cross-Domain Study in Italian}


\author*[1]{\fnm{Serena} \sur{Auriemma}}\email{serena.auriemma@phd.unipi.it}

\author[1]{\fnm{Martina} \sur{Miliani}}\email{martina.miliani@fileli.unipi.it}

\author[2]{\fnm{Mauro} \sur{Madeddu}}\email{mauro.madeddu@phd.unipi.it}

\author[1,2]{\fnm{Alessandro} \sur{Bondielli}}\email{alessandro.bondielli@unipi.it}

\author[2]{\fnm{Lucia} \sur{Passaro}}\email{lucia.passaro@unipi.it}

\author[1]{\fnm{Alessandro} \sur{Lenci}}\email{alessandro.lenci@.unipi.it}

\affil*[1]{\orgdiv{CoLing Lab, Department of Philology, Literature and Linguistics}, \orgname{University of Pisa}, \orgaddress{\street{36 Santa Maria Street}, \city{Pisa}, \postcode{56126}, \country{Italy}}}

\affil[2]{\orgdiv{Department of Computer Science}, \orgname{University of Pisa}, \orgaddress{\street{3 Largo Bruno Pontecorvo}, \city{Pisa}, \postcode{56127}, \country{Italy}}}



\abstract{
Addressing the challenge of limited annotated data in specialized fields and low-resource languages is crucial for the effective use of Language Models (LMs). While most Large Language Models (LLMs) are trained on general-purpose English corpora, there is a notable gap in models specifically tailored for Italian, particularly for technical and bureaucratic jargon. This paper explores the feasibility of employing smaller, domain-specific encoder LMs alongside prompting techniques to enhance performance in these specialized contexts.
Our study concentrates on the Italian bureaucratic and legal language, experimenting with both general-purpose and further pre-trained encoder-only models. We evaluated the models on downstream tasks such as document classification and entity typing and conducted intrinsic evaluations using Pseudo-Log-Likelihood.
The results indicate that while further pre-trained models may show diminished robustness in general knowledge, they exhibit superior adaptability for domain-specific tasks, even in a zero-shot setting. Furthermore, the application of calibration techniques and in-domain verbalizers significantly enhances the efficacy of encoder models. These domain-specialized models prove to be particularly advantageous in scenarios where in-domain resources or expertise are scarce.
In conclusion, our findings offer new insights into the use of Italian models in specialized contexts, which may have a significant impact on both research and industrial applications in the digital transformation era.
}






\maketitle

\section{Introduction}\label{intro}

Pre-trained LMs have had a significant impact on Natural Language Processing (NLP), with the ``pre-train and fine-tune'' paradigm rapidly becoming the predominant approach to apply effective models on a wide variety of downstream tasks \cite[\textit{inter alia}]{vaswani2017attention,howard-ruder-2018-universal,devlin-etal-2019-bert}. 
However, one of the main concerns when working with LMs is the paucity of annotated data, especially for specific domains or low-resource languages, required to fine-tune the additional classification layer on top of these models for downstream tasks, such as classification. Recently, prompt-based tuning has started to affirm as a promising way to perform similar tasks, significantly reducing the need for annotated data. 
This approach has been proven to be very effective with Large Language Models (LLMs) \cite{Brown2020Language}. However, it is often the case that LLMs are not available for low-resource languages, and that their performance drastically decreases when they are challenged on specific domains. 
Moreover, in the Digital Transformation era, businesses frequently need to integrate artificial intelligence systems into their application ecosystems. This requires them to utilize specialized, publicly available models while also employing effective methods to leverage these models in scenarios where annotated language resources are unavailable, thereby operating in a zero-shot mode.

Hence, we decided to evaluate two smaller domain-specific encoder models: BureauBERTo \cite{auriemma2023bureauberto}, a LM further pre-trained on Italian bureaucratic texts (i.e., administrative acts, banking and insurance documents), and Italian Legal BERT \cite{licari2022italian} (henceforth referred to as Ita-Legal-BERT), a LM adapted to the Italian legal domain, on various classification tasks on domain-specific data exploiting a prompt-based technique in a zero-shot scenario. Additionally, we compared the performance of both models with that of a generic Italian model, UmBERTo.\footnote{\url{https://huggingface.co/Musixmatch/umberto-commoncrawl-cased-v1}} 

Since BureauBERTo has proved to be particularly accurate in a fill-mask task \cite{auriemma2023bureauberto},\footnote{See Appendix \ref{ap:fillmask} for the plot of the model results in the fill-mask task.} in which the model had to predict both random and in-domain masked words, we sought to further inspect the domain lexical knowledge acquired by this model during the domain adaptation. We aimed at leveraging this knowledge to implement two classification tasks in the Public Administration (PA) domain, modeled as prompt-based classification. Specifically, we challenged the model to predict both the topics of PA texts, and the type of generic and PA-related named entities occurring in sentences extracted from administrative documents. Additionally, we aimed to evaluate BureauBERTo on a language variety kindred to its domain of specialization, namely Italian legal language. Furthermore, we wanted to investigate whether there are differences in the performance of specialized models obtained through additional pre-training with and without a preliminary vocabulary expansion with domain words (see Section \ref{sec:models}), particularly when the tasks are modeled as prompt based classification and heavily rely on the models' vocabulary. Therefore, we tested two additional datasets for the classification of topics and domain entities related to the legal domain, and we compared the performance of BureauBERTo with models with no domain tokens added to the vocabulary, such as the specialized Ita-Legal-BERT and the domain-agnostic model UmBERTo.

Using encoder-only models with prompting requires transforming the original classification task into a cloze task. In this approach, the model evaluates each classification class by assigning a probability to a set of label words that fill the masked position within the prompt. This set of label words and their corresponding classes is known as the \textit{verbalizer}.

The prompting experiments were carried out across multiple settings, involving the use of three distinct types of verbalizers: i.) a base verbalizer, ii.) a manually constructed verbalizer, and iii.) a verbalizer automatically generated by each model. With these three resources, we tried to investigate how domain-related word labels affect each model's performance in different classification tasks.
Additionally, we implemented two forms of model calibration to mitigate the models' sensitivity to both the prompt template and the label words encompassed within the verbalizers \cite{zhao2021calibrate}.
Finally, we computed  Pseudo-Log-Likelihood scores (PLLs) \cite{salazar2019masked} for all models on representative samples extracted from the classification datasets in both the PA and legal domain, as well as on a generic corpus used as a baseline. This approach aims to measure the linguistic competence of the models in the two target domains and to assess its correlation with their performance on the domain-specific tasks.

To the best of our knowledge, this is the first work to thoroughly explore zero-shot classification using encoder-only models of relatively small size and focusing on the PA and legal domains for the Italian language. The main contributions of this work are:
\begin{itemize}
    \item An evaluation of two specialized models, BureauBERTo and Ita-Legal-BERT, in zero-shot classification tasks using prompt-based technique.
    \item An analysis of the impact of domain-specific pre-training and vocabulary expansion on model performance.
    \item The implementation of multiple verbalizers and calibration methods to inspect models sensitivity to prompts and word labels.
    \item The computation of PLLs to assess the relationship between models linguistic competence and task performance.
\end{itemize} 

The rest of this paper is organized as follows. Section \ref{sec:relworks} reviews related work and background information on LMs and prompt-tuning. Section \ref{sec:expset} details the experimental setup, including the models, datasets, calibration methods, verbalizers, and prompt-template used. Section \ref{sec:results} presents the results of our experiments. Finally, Section \ref{sec:conclusion} discusses the findings and their implications, and outlines directions for future research.

\section{Related work} \label{sec:relworks}

\subsection{Prompt-tuning}

Pre-trained LMs have proven to be effective in NLP tasks related to specific domains, whether they were trained from scratch \cite{beltagy2019scibert,gu2021domain}, or further pre-trained on domain data \cite{lee2020biobert,chalkidis2020legal,licari2022italian} with a Masked Language Modeling (MLM).  
More recently, the MLM training objective has been leveraged to solve various NLP tasks reformulated as a sort of cloze task, allowing the LM to directly solve it without any or with very few labeled examples. One of the first works 
in this direction was proposed by \cite{radford2019language}, who performed 
zero-shot learning using pre-trained LMs without fine-tuning on a dataset of training examples.  
Within similar conditions, but using the larger GPT-3, near state-of-the-art results have been achieved \cite{Brown2020Language} for some SuperGLUE tasks \cite{wang2019superglue}.  
Competitive performance with GPT-3 has been attained using much smaller models like the 220M parameters ALBERT, by performing some gradient-based fine-tuning of the model using the labeled examples on a cloze task \cite{schick-schutze-2021-just}. Since then, prompt-based learning has gained attention as a simple way to perform, among other tasks, zero-shot classification \cite{puri2019zero}.  
However, it is essential to note that the performance of prompt-based learning techniques scales with model size \cite{lester2021power}.
Consequently, general-purpose LLMs with billions of parameters are typically used in prompt-learning experiments, even for specialized domains such as the legal one \cite{yu2022legal}. In contrast, for the biomedical and clinical domains, smaller specialized models like BioBERT \cite{lee2020biobert} and Clinical BERT \cite{alsentzer2019publicly} outperformed GPT-2 and T5 in a few-shot prompt based classification of medical texts \cite{sivarajkumar2022healthprompt}. The authors hypothesize that the advantage of the BERT-based models is possibly due both to their domain adaptation and to their bidirectional MLM training objective, which is more similar to the prompt template format than those of auto-regressive and sequence-to-sequence models like GPT-2 and T5. A similar finding was reported by \cite{moradi2021gpt} even for the much larger GPT-3 over BioBERT. Nevertheless, these approaches are constrained by the model input size, which limits the length of the conditioning input context and can significantly affect performance \cite{moradi2021gpt}.

Although prompt-based classification with specialized models has been explored for the medical and clinical domains, to the best of our knowledge, this is the first work that focuses on applying prompts to the Italian administrative and legal language in a zero-shot classification scenario. 

\subsection{Mapping labels through the verbalizer}

To perform prompt-based classification with encoder-only LMs, a specific template is required to reformulate the original classification task as a cloze-task, where the text to be classified is fed to the model followed by a prompt sentence, such as ``This \xml{text} is about [MASK]''. In this way, the model has to predict the probability that a certain word is filled in the ``[MASK]'' token. The mapping from the label candidate word to a specific class is gained through the \textit{verbalizer} \cite{schick-schutze-2021-just}, which represents the original class names as a set of label words. The choice of the verbalizer greatly influences the model performance \cite{gao-etal-2021-making}. 

Usually, verbalizers are manually constructed by associating each class with one or few word labels that capture the main content of the class texts and on which the model relies to perform the classification \cite{schick-schutze-2021-just,schick-schutze-2021-exploiting}. However, selecting appropriate words is not trivial because the probability of the chosen label words being the ``correct'' token in the [MASK] position within the prompt determines the model's accuracy in the task. To reduce the human effort in handcrafting the verbalizer,  another approach automatically finds the label mapping using discrete search on a small amount of training data \cite{schick2020automatically}. Nevertheless, automatically defined verbalizers still perform worse than manual ones, especially in zero-shot settings \cite{gao-etal-2021-making,hu2021knowledgeable}. 

When working with specific domains, manually defining a verbalizer requires both domain expertise and an understanding of the specialized model linguistic competence. \cite{hu2021knowledgeable} proposed the \textit{knowleadgeable prompt-tuning} (KPT) method to address this issue. KPT expands the verbalizer word labels by leveraging an external knowledge base, such as a knowledge graph resource like Related Words.\footnote{\url{https://relatedwords.org}} Then, the expanded verbalizer can be refined using the LM itself by excluding infrequent words to which the model assign lower probabilities, out-of-vocabulary tokens, words that may positively contribute to multiple classes, and by leveraging contextual calibration (see Section \ref{relwork-calibration}).

To assess the impact of the verbalizer choice on task performance, we tested the models using three types of verbalizer for each task: a base verbalizer, a manual verbalizer, and a knowledgeable verbalizer. Due to the unavailability of resources like Related Words for our specific domains and the Italian language, we constructed the external knowledge base using the LOTClass method \cite{meng2020text}, following the approach suggested by \cite{hu2021knowledgeable}.  

Further details on how we constructed the external knowledge base and all the verbalizers are reported in Section 
\ref{sec:expsetverb}.   

\subsection{Model stabilization through calibration} \label{relwork-calibration}

A notable challenge in prompt-based approaches lies in their sensitivity to variations in prompt templates and verbalizers  \cite{hu2021lora,lu2022fantastically,trautmann2022legal}. 
This issue is particularly evident in the so-called "in-context" learning where the model learns how to solve a task given some human-designed instruction and a few demonstrations provided in the prompt, without requiring any parameters update. According to \cite{zhao2021calibrate}, the instability of results in few-shot learning may be attributed to three types of biases resulting from the prompt:
\begin{itemize}
\item the \textit{recency bias}, i.e., the model tends to repeat the answer that appears most recently, typically towards the end of the prompt;
\item the \textit{majority label bias}, which occurs when there is a class imbalance in a text classification prompt, such as having more Positive than Negative sentiment examples;
\item the \textit{common token bias} according to which the model is more prone to output tokens that are frequent in its pre-training distribution.
\end{itemize} 

The last bias is particularly relevant for text classification tasks, even in zero-shot settings, because the model makes predictions by choosing label words associated with specific classes. As certain label words appear more frequently in its pre-training corpus, the model is inherently biased toward predicting certain classes.
To mitigate these biases, various calibration approaches have been proposed.

The \textbf{contextual calibration} (CC) was first introduced by \cite{zhao2021calibrate}. It relies on the assumption that a model's bias towards a certain class can be estimated by feeding the model a content-free input string, such as "N/A", an empty string, or gibberish tokens. Ideally, given a content-free string, the model would assign an equal score to all classes. In reality, due to the biases described above, the model tends to assign a higher probability to a specific class. This type of error is defined as \textit{contextual}

because a different template, training example, permutation, or word labels would likely lead to a different output, given the same content-free input. Then, one can fit the calibration parameters derived from the prediction on the content-free input to be uniform across classes. Based on a similar approach, the \textbf{domain-context calibration} \cite{fei2023mitigating} involves calibrating the model using a sequence of words randomly sampled from the unlabelled evaluation dataset of the task instead of an empty string.

However, according to \cite{zhou2023batch} both contextual and domain calibrations may be suboptimal for certain kinds of tasks, and might even introduce additional biases. For example, in a textual entailment task, having the same content-free tokens (e.g., "N/A") in both the question and the answer could bias the model to predict the class \textit{entailment}, due to the identical forms of the two input texts. Using randomly sampled words as input strings to calibrate the model, especially in conjunction with in-context demonstrations, could cause spurious relations among them, further influencing the model output. Therefore, they proposed a \textbf{batch-calibration} (BC) approach that, instead of relying on content-free strings to estimate the contextual bias, followed a context-based approach using a small batch of unlabelled text examples from each class. This latter procedure is particularly pertinent to zero-shot classification scenarios.

Given the negative impact of these biases on LMs performance in prompt-based classification tasks, we addressed this issue by calibrating the models following the \textit{contextual} and \textit{batch calibration} approaches. Despite the limitations highlighted by \cite{zhou2023batch}, we also tested contextual calibration, as this approach is particularly useful when data related to a specific domain is lacking. We did not use the \textit{domain-context calibration} because our work focuses primarily on specialized models and specific domains where domain terms often occur in the input texts and as word labels: even by using random words, we would have affected the models' behavior. Further details about the model calibration we applied in our tasks are reported in Section \ref{sec:expsetcal}.  

\section{Experimental settings} \label{sec:expset}

To conduct our experiments we used OpenPrompt\footnote{\url{https://github.com/thunlp/OpenPrompt}}, an open-source Python framework for prompt-learning. This framework also provides methods to experiment with different types of verbalizers, such as manual and knowledgeable verbalizers, and to calibrate models. 
We conducted our experiments on an NVIDIA A100 GPU.

\subsection{Models} \label{sec:models}
For our experiments, we compared the performance of three encoder-only models, namely UmBERTo, Ita-Legal-BERT and BureauBERTo. 

UmBERTo\footnote{\url{https://github.com/musixmatchresearch/umberto}} is a RoBERTa-based \cite{liu2019roberta} language model trained on the Italian section of the OSCAR corpus,\footnote{\url{https://oscar-corpus.com}} that has been shown to perform well on administrative data \cite{auriemma2022evaluating} compared to other generic PLMs of the same size (110M parameters).

Ita-Legal-BERT\footnote{\url{https://huggingface.co/dlicari/Italian-Legal-BERT}} was released by \cite{licari2022italian}, who further pre-trained Bert-Ita\footnote{\url{https://huggingface.co/dbmdz/bert-base-italian-xxl-cased}} on 235,000 legal documents (civil and criminal) from the National Case Law Archive (Archivio Giurisprudenziale Nazionale). Ita-Legal-BERT has been successfully employed for NER on civil judgments and to classify sentences extracted from judgments. 

BureauBERTo\footnote{\url{https://huggingface.co/colinglab/BureauBERTo}} \cite{auriemma2023bureauberto} is a domain-adapted model obtained by further pre-training UmBERTo on Italian administrative, banking, and insurance documents. To enhance the model's ability to understand domain-specific language, the vocabulary of BureauBERTo has been enriched by incorporating 8,305 new domain tokens into the original 32,000-token vocabulary of UmBERTo, resulting in a tokenizer with 40,305 tokens. This expansion increased the model's parameter count from 110 million to 117 million.

\subsection{Datasets} \label{sec:datasets}
We tested the models on two different domains, PA and legal, and we used four different datasets, two for each classification task.

\subsubsection{PA} \label{sec:datapa}
For the \textbf{prompt document classification}, we used a subset of the ATTO corpus \cite{auriemma2022evaluating}, which is a collection of short administrative documents annotated with labels denoting topics. We filtered this dataset keeping only those instances (2,812) that were annotated with a single topic label (see Table \ref{tab:LabeldistrATTO} in Appendix \ref{ap:datasets} for the distribution of the labels in the dataset). We split the dataset, using 90\% for testing the models and 10\% for performing batch calibration and creating the knowledgeable verbalizer.

For the \textbf{prompt entity typing} task, we used the PA-corpus of \cite{passaro2017informed}, a collection of 460 PA documents with token-level annotations of Named Entities denoting both general entities, such as persons (\textsc{PER}; 3,706), locations (\textsc{LOC}; 4,498), organizations (\textsc{ORG}; 3,594), and domain-specific entities, like legislative norms (\textsc{LAW};  5,217), acts (\textsc{ACT}; 2,240), and PA-related organizations (\textsc{OPA}; 2,074). As a test set, we adopted the same subset as the one reported by \cite{auriemma2022evaluating} corresponding to the 10\% of the dataset. To build the knowledgable verbalizer we used the training set, i.e., a subsample corresponding to the 70\% of the dataset. We extract 180 sentences from this subset to apply batch calibration.

\subsubsection{Legal} \label{sec:datalegal}

For the \textbf{prompt-document classification} in the legal domain, we used a dataset comprising 3,190 Italian civil judgments annotated with various topics (see Table \ref{tab:LabeldistrLegal} in Appendix \ref{ap:datasets} for the complete list). 
Italian civil judgments follow a quasi-standardized structure, which includes sections such as the heading, the parties involved, the subject matter, the factual background, the legal reasoning, and the judge's decision. To classify judgments based on their main topics, we provided the models with only the factual background sections of the documents. This section presents an account of the facts of the case, making it the most informative part regarding the judgment topics. Additionally, the models' maximum input length of 512 tokens, necessitated focusing on the most relevant part of the judgments. As for the PA domain, we divided the dataset into 90\% for testing and 10\% for constructing the knowledgeable verbalizer and performing batch calibration.

As for the \textbf{entity typing} task, we used a dataset composed of 57 Italian civil judgments. The dataset counts 4 different labels: \textsc{JDG} for judges (332 instances), \textsc{LWY} for lawyers (302 instances), \textsc{LAW} for references to laws (976), and \textsc{RUL} for references to other judgments (440). We used 90\% of the sentences to create the knowledgeable verbalizer. From this subset, we sampled 120 sentences to apply batch calibration. We used the remaining 10\% to test the models.

\subsection{Verbalizers} \label{sec:expsetverb} 

We conducted our prompt-based classification experiments in the PA domain using three types of verbalizers, namely a base, a manual and a knowledgeable one.  For the legal domain, we focused only on the base and the knowledgeable verbalizers, as we lacked the domain expertise necessary to depict civil judgements topics at a finer-grained level.
The aim of this comparison was to better understand the correlation between the lexical knowledge of LMs and the use of domain-related terms as the set of word labels encompassed within the verbalizers. 

The \textbf{Base-verbalizer} simply uses the Italian name of the classification classes reported in the datasets as label words (e.g., \textit{Ambiente - ``Environment''} is the label word for the class \textsc{Ambiente - ``Environment''} in the PA document classification task.)

The \textbf{Manual-verbalizer} is a verbalizer that we constructed by adding some synonyms of the class name and some related PA terms as label words for each class, to  better depict the documents content and the entity types (in this case the label words for the class \textsc{Ambiente - ``Environment''} are: \textit{ ambiente - ``environment'', natura - ``nature'', territorio - ``territory'', flora - ``flora'', etc.}).

The complete list of the label words used in both base and manual verbalizers in the PA domain is shown in Table \ref{docclass-verbalizer} for document classification,\footnote{See Appendix \ref{ap:labelwords} for the English translation of the label words for document classification.} and in Table \ref{et-verbalizer} for entity typing. For the base verbalizers in the legal domain, see Tables \ref{docclass-verbalizer-legal} and \ref{et-verbalizer-legal} in Appendix \ref{ap:labelwords}.

\begin{table*}[!ht]
\caption{ The label words used in the base and manual verbalizers for the prompt-based document classification experiments in the PA domain.}\label{docclass-verbalizer}
\centering \small
\setlength{\tabcolsep}{5pt}
\renewcommand\tabularxcolumn[1]{m{#1}}
\begin{tabularx}{1\textwidth}{@{}llX@{}}
\makecell{\textbf{Class}} & \makecell{\textbf{Base Verbalizer}} & \textbf{Manual Verbalizer}\\
\midrule
\makecell{\textsc{Ambiente}} & \makecell{\textit{ambiente}} & \RaggedRight{\textit{ambiente, natura, territorio, flora, fauna, animali, clima, inquinamento, rifiuti, igiene, caccia, pesca, verde, ecologia, agricoltura, acque}}\\
\midrule
\makecell{\textsc{Avvocatura}} & \makecell{\textit{avvocatura}} & \RaggedRight{\textit{avvocatura, avvocati, giustizia, legale, ricorso, giudici, Tribunale, Corte, Appello, Assise, notifica, atti, Albo, Pretorio, protocollo}}\\
\midrule
\makecell{\textsc{Bandi e contratti}} & \makecell{\textit{bandi},\\ \textit{contratti}} & \RaggedRight{\textit{bandi, contratti, bando, contratto, gara, appalto, assunzione, liquidazione}}\\
\midrule
\makecell{\textsc{Commercio e}\\\textsc{attività}\\\textsc{economiche}} & \makecell{\textit{commercio,}\\ \textit{attività,}\\ \textit{economiche}} & \RaggedRight{\textit{commercio, economia, attività, economica, beni, commerciare, vendite, acquisti, commercianti, confesercenti}}\\
\midrule
\makecell{\textsc{Cultura,}\\\textsc{turismo,}\\\textsc{ e sport}} & \makecell{\textit{cultura,}\\ \textit{turismo,}\\ \textit{sport}} & \RaggedRight{\textit{cultura, turismo, sport, culturale, turisti, musei, arte, cinema, vacanze, spettacolo, scuola, manifestazioni}}\\
\midrule
\makecell{\textsc{Demografico}} & \makecell{\textit{demografico}} & \RaggedRight{\textit{demografico, popolazione, abitanti, residenti, censimento, anagrafe, residenza, domicilio, cittadinanza, leva}}\\
\midrule
\makecell{\textsc{Edilizia}} & \makecell{\textit{edilizia}} & \RaggedRight{\textit{edilizia, costruzioni, cantiere, ristrutturazione, planimetrie, residenziale}}\\
\midrule
\makecell{\textsc{Personale}} & \makecell{\textit{personale}} & \RaggedRight{\textit{personale, risorse, umane, assunzioni, lavoro, part-time}}\\
\midrule
\makecell{\textsc{Pubblica istruzione}} & \makecell{\textit{istruzione}} & \RaggedRight{\textit{istruzione, istituto, scolatisco, scuola, insegnante, formazione, educazione}}\\
\midrule
\makecell{\textsc{Servizi}\\\textsc{informativi}} & \makecell{\textit{servizi,}\\ \textit{informazioni}} & \RaggedRight{\textit{servizi, informazioni, informativi}}\\
\midrule
\makecell{\textsc{Servizio}\\\textsc{finanziario}} & \makecell{\textit{finanza}} & \RaggedRight{\textit{finanza, euro, finanziario, contabilità, contabile, copertura, rimborsi, pagamenti, versamenti, bilancio, spese, sanzioni, multe, tributi, retribuzioni, emolumenti}}\\
\midrule
\makecell{\textsc{Sociale}} & \makecell{\textit{sociale}} & \RaggedRight{\textit{sociale, leva, militare, disabili, protezione, civile, invalidi}}\\
\midrule
\makecell{\textsc{Urbanistica}} & \makecell{\textit{urbanistica}} & \RaggedRight{\textit{urbanistica, trasporti, trasporto, traffico, circolazione, veicoli, viabilità, viaria}}\\
\bottomrule
\end{tabularx}
\end{table*}

\begin{table*}[!ht]
\centering \small
\setlength{\tabcolsep}{3pt}
\caption{\label{et-verbalizer} The label words used in the base and manual verbalizers for the prompt entity typing experiments in the PA domain.
}
\renewcommand\tabularxcolumn[1]{m{#1}}
\begin{tabularx}{\textwidth}{@{}clX@{}}
\toprule
 \textbf{Class} & \textbf{Base Verbalizer} & \textbf{Manual Verbalizer} \\

\midrule
\textsc{PER} & \textit{persona} & \textit{persona (person), generalità (particulars), nominativo (name)} \\
\midrule
\textsc{LOC} & \textit{luogo}  & \textit{luogo (place), località (locality)}\\
\midrule
\textsc{ORG} & \textit{organizzazione} & \RaggedRight{\textit{organizzazione (organization), azienda (firm), società (corporation), associazione (association), compagnia (company)}} \\
\midrule
\textsc{LAW} & \textit{legge} & \RaggedRight{\textit{legge (law), norma (rule), decreto (decree), legislativo (legislative)}}\\
\midrule
\textsc{ACT} & \textit{atto} & \RaggedRight{\textit{atto (act), delibera (resolution), determina (decision), deliberazione (deliberation),  regolamento (regulation)}} \\
\midrule
\textsc{OPA} & \textit{ufficio}  & \textit{ufficio (office)} \\

\bottomrule
\end{tabularx}

\end{table*}

Finally, the \textbf{Knowledgeable verbalizer} was created following the Label-Name Only Text Classification (LOTClass) method by \cite{meng2020text} to build the external knowledge-base, which was then refined using the approach of \cite{hu2021knowledgeable}. The LOTClass method enables constructing a category vocabulary for each class that contains semantically related words to the label name of the class by leveraging the model vocabulary via MLM.  We applied this method to additional samples from each dataset not used for testing the models.
We can summarize the applied methodology in the four following steps. 

\begin{enumerate}
    \item \textbf{Class name occurrences.} We searched for occurrences of each class name, a few synonyms, and its grammatical variants (e.g., plural/singular), exclusively in texts belonging to that class, and masked these words using the rest of the sentence as context. Specifically, we identified 1,226 sentences containing class names in the PA document classification dataset, and 236 in the legal dataset. For the entity typing task, we found 506 sentences containing entity names in the legal dataset and 253 in the PA domain. 
    \item \textbf{Class Vocabulary.} We then performed MLM with the three models separately to generate 50 candidate words for each occurrence. This process yielded a class vocabulary (CV) with the top 100 candidate words for each class, ranking candidates by their frequency as fillers for the masked word and discarding stopwords and words appearing in multiple classes.
    \item \textbf{Category-informative filtering.} To ensure that we masked only category-informative occurrences of a word (i.e., words whose contextual meaning is strongly associated to the class texts content), we filtered out sentences where fewer than 20 out of the 50 requested candidate words did not belong to the previously constructed CV, and repeated the MLM only on the remaining sentences. At the end of this process, we obtained a sample of approximately 100 in context relate words to each class. It is important to note that this approach is model-dependent, as each model produces candidates based on its own vocabulary.These candidates were then integrated in 12 different external knowledge bases, one for each domain, task, and model.
    \item \textbf{Knowledgeable verbalizer refinement.} The knowledgeable verbalizer class within OpenPrompt framework allowed us to refine the word labels associated with the various classes using the following methods: \textit{frequency refinement}, words to which the model tends to assign probabilities lower than a threshold were removed; \textit{relevance refinement}, filter out words that may possibly contribute positively for multiple classes;  discarding sub-tokens; and contextual calibration. Given the considerable number of words in our external knowledge base (approximately 100 words per class), we applied all the available refinement methods. 
\end{enumerate}

Examples of words from the final knowledgeable verbalizers of Ita-Legal-BERT, BureauBERTo, and UmBERTo are listed in Table \ref{tab:knowverbaDC} for the document classification tasks and in Table \ref{tab:knowverbaET} for the entity typing tasks.\footnote{See Appendix \ref{ap:labelwords} for the English translation of the label words used for both document classification and entity typing tasks.}

\begin{table}[h]
\centering
\caption{Some of the label words in the knowledgeable verbalizers of Ita-Legal-BERT, BureauBERTo, and UmBERTo constructed for the experiments on prompt document classification.}
\label{tab:knowverbaDC}
\begin{tabular}{lllccc}
\toprule
\multirow{2}{*}{\textbf{Domain}} & \multirow{2}{*}{\textbf{Class}} & \multicolumn{3}{c}{\textbf{Knowledgeable Verbalizer}} \\
\cmidrule(lr){3-5}
& & Ita-Legal-BERT & BureauBERTo & UmBERTo \\
\midrule
\multirow{2}{*}{\textbf{PA}} & \begin{tabular}[c]{@{}c@{}} \textsc{Commercio e} \\ \textsc{attività} \\ \textsc{economiche}\end{tabular} & \begin{tabular}[c]{@{}c@{}} \textit{quote,} \\ \textit{prestazioni,} \\ \textit{obbligazioni,} \\ \textit{vendite,} \\ \textit{imposte,} \\ \textit{transazioni, ...}\end{tabular} & \begin{tabular}[c]{@{}c@{}} \textit{attività,} \\ \textit{professioni,} \\ \textit{licenze,} \\ \textit{imprese,} \\ \textit{Associazioni,} \\ \textit{negozi, ...}\end{tabular} & \begin{tabular}[c]{@{}c@{}} \textit{attività,} \\ \textit{utilità,} \\ \textit{esercizi,} \\ \textit{opere,} \\ \textit{benefici,} \\ \textit{attrezzature, ...}\end{tabular} \\
\\
& \textsc{Avvocatura} & \begin{tabular}[c]{@{}c@{}} \textit{consiglio,} \\ \textit{avvocato,} \\ \textit{giudizi,} \\ \textit{tribunale,} \\ \textit{difensori,} \\ \textit{Giudice,  ...}\end{tabular} & \begin{tabular}[c]{@{}c@{}} \textit{decreto,} \\ \textit{procedura,} \\ \textit{decisione,} \\ \textit{archiviazione,} \\ \textit{atto,} \\ \textit{sentenza, ...}\end{tabular} & \begin{tabular}[c]{@{}c@{}} \textit{decreto,} \\ \textit{delibera,} \\ \textit{delega,} \\ \textit{definizione,} \\ \textit{delegato,} \\ \textit{deliberazione, ...}\end{tabular} \\
\midrule

\multirow{2}{*}{\textbf{Legal}} & \begin{tabular}[c]{@{}c@{}} \textsc{Polizia} \\ \textsc{locale} \end{tabular} & \begin{tabular}[c]{@{}c@{}} \textit{vigilanza,} \\ \textit{sorveglianza,} \\ \textit{circolazione,} \\ \textit{strada,} \\ \textit{guardia,} \\ \textit{soccorso, ...}\end{tabular} & \begin{tabular}[c]{@{}c@{}} \textit{verifica,} \\ \textit{disciplina,} \\ \textit{vigilanza,} \\ \textit{Prefettura,} \\ \textit{Guardia,} \\ \textit{sicurezza, ...}\end{tabular} & \begin{tabular}[c]{@{}c@{}} \textit{prevenzione,} \\ \textit{vigilanza,} \\ \textit{verifica,} \\ \textit{guardia,} \\ \textit{sorveglianza,} \\ \textit{commissione, ...}\end{tabular} \\
\\
& \begin{tabular}[c]{@{}c@{}} \textsc{Appalti e} \\ \textsc{contratti} \end{tabular} & \begin{tabular}[c]{@{}c@{}} \textit{gara,} \\ \textit{contratto,} \\ \textit{convenzione,} \\ \textit{bando,} \\ \textit{prezzo,} \\ \textit{vendita, ...}\end{tabular} & \begin{tabular}[c]{@{}c@{}} \textit{modulo,} \\ \textit{contratto,} \\ \textit{rinnovo,} \\ \textit{accordo,} \\ \textit{compromesso,} \\ \textit{preliminare, ...}\end{tabular} & \begin{tabular}[c]{@{}c@{}} \textit{disciplinare,} \\ \textit{codice,} \\ \textit{testo,} \\ \textit{avviso,} \\ \textit{trattato,} \\ \textit{decreto, ...}\end{tabular} \\

\bottomrule
\end{tabular}
\end{table}

\begin{table}[h]
\centering
\caption{Some of the label words in the knowledgeable verbalizers of Ita-Legal-BERT, BureauBERTo, and UmBERTo constructed for the experiments on prompt entity typing.}
\label{tab:knowverbaET}
\begin{tabular}{lllccc}
\toprule
\multirow{2}{*}{\textbf{Domain}} & \multirow{2}{*}{\textbf{Class}} & \multicolumn{3}{c}{\textbf{Knowledgeable Verbalizer}} \\
\cmidrule(lr){3-5}
& & Ita-Legal-BERT & BureauBERTo & UmBERTo \\
\midrule 
\multirow{2}{*}{\textbf{PA}} & \begin{tabular}[c]{@{}c@{}} \textsc{law} \end{tabular} & \begin{tabular}[c]{@{}c@{}} \textit{Regolamento,} \\ \textit{direttiva,} \\ \textit{normativa,} \\ \textit{bando,} \\ \textit{norma,} \\ \textit{riforma, ...}\end{tabular} & \begin{tabular}[c]{@{}c@{}} \textit{Regolamento,} \\ \textit{Decreto,} \\ \textit{Legge,} \\ \textit{deliberazione,} \\ \textit{norma,} \\ \textit{decreti, ...}\end{tabular} & \begin{tabular}[c]{@{}c@{}} \textit{contratto,} \\ \textit{certificato,} \\ \textit{Protocollo,} \\ \textit{codice,} \\ \textit{normativa,} \\ \textit{titolo, ...}\end{tabular} \\
\\
& \textsc{opa} & \begin{tabular}[c]{@{}c@{}} \textit{ufficio,} \\ \textit{reparto,} \\ \textit{Responsabile,} \\ \textit{coordinatore,} \\ \textit{dipartimento,} \\ \textit{Istituto, ...}\end{tabular} & \begin{tabular}[c]{@{}c@{}} \textit{servizio,} \\ \textit{settore,} \\ \textit{presidio,} \\ \textit{reparto,} \\ \textit{Dipartimento,} \\ \textit{istituto, ...}\end{tabular} & \begin{tabular}[c]{@{}c@{}} \textit{campo,} \\ \textit{servizio,} \\ \textit{operatore,} \\ \textit{ramo,} \\ \textit{reparto,} \\ \textit{settore, ...}\end{tabular} \\
\midrule

\multirow{2}{*}{\textbf{Legal}} & \begin{tabular}[c]{@{}c@{}} \textsc{lwy} \end{tabular} & \begin{tabular}[c]{@{}c@{}} \textit{difensore,} \\ \textit{amministratore,} \\ \textit{Av,} \\ \textit{attore,} \\ \textit{attrice,} \\ \textit{assistito, ...}\end{tabular} & \begin{tabular}[c]{@{}c@{}} \textit{Avv,} \\ \textit{amministratore,} \\ \textit{ufficio,} \\ \textit{agente,} \\ \textit{difensore,} \\ \textit{avvocati, ...}\end{tabular} & \begin{tabular}[c]{@{}c@{}} \textit{impiegato,} \\ \textit{operatore,} \\ \textit{impresa,} \\ \textit{agenzia,} \\ \textit{legale,} \\ \textit{Avv, ...}\end{tabular} \\
\\
& \begin{tabular}[c]{@{}c@{}} \textsc{rul} \end{tabular} & \begin{tabular}[c]{@{}c@{}} \textit{arresto,} \\ \textit{pronuncia,} \\ \textit{dispositivo,} \\ \textit{verbale,} \\ \textit{sentenze,} \\ \textit{condanna, ...}\end{tabular} & \begin{tabular}[c]{@{}c@{}} \textit{determinazione,} \\ \textit{dottrina,} \\ \textit{giudizio,} \\ \textit{Corte,} \\ \textit{causa,} \\ \textit{Giurisprudenza, ...}\end{tabular} & \begin{tabular}[c]{@{}c@{}} \textit{domanda,} \\ \textit{giudizio,} \\ \textit{decisioni,} \\ \textit{verbale,} \\ \textit{citazione,} \\ \textit{ordinanza, ...}\end{tabular} \\

\bottomrule
\end{tabular}
\end{table}

\subsection{Calibration} 
\label{sec:expsetcal}

To stabilize the models' results, we calibrated them for both entity typing and document classification tasks in the PA and legal domains. Calibration was performed twice for each task using the Contextual Calibration (CC) and Batch Calibration (BC) approaches to evaluate how different calibration methods affect the performance of generic vs. specialized models, and with different verbalizers. In the CC settings, we used an empty string as content-free input. In the BC approach, we calibrated the models using a stratified sample of unlabeled texts extracted from the dataset that we did not use for predictions. Specifically, for document classification in the PA and legal domains, we used samples of 200 documents, while for the entity typing task, we used 180 sentences for the PA domain and 120 for the legal domain.

\subsection{Prompt-based classification}

We performed the prompt-based classification across different settings. For each domain, we performed CC and BC on all three models, and compared the performance with that obtained without any calibration.

We tested the models using: a base verbalizer, where each class is linked to one or few label words that correspond to the names of the classes in the original dataset annotations described in Section \ref{sec:datasets} (e.g., the one from the ATTO corpus and the InformedPA corpus for the PA domain); a manual verbalizer, enriched by a collection of domain terms manually selected as representative for the classification labels; and a knowledgeable verbalizer, specific to each model. 
For the knowledgeable verbalizer (See Section \ref{sec:expsetverb}), we created three different versions for each domain and task, using the two specialized models, Ita-Legal-BERT and BureauBERTo, and the generic model UmBERTo. We tested the models using only their respective verbalizers.

For the experiments in the legal domain, we tested the models using only the base and the knowledgeable verbalizers.

\subsubsection{Entity Typing} \label{sec:task-et}

We modeled the NER task introduced by~\cite{passaro2017informed} as an entity typing task. Entity typing can be considered a subtask of NER and focuses on entity classification. In other words, systems assign a label to an already extracted entity. This task is often formulated to challenge systems at retrieving sub-categories organized in a hierarchical structure (e.g., an entity corresponding to a person may be specified as director, major, lawyer, etc.) As in~\cite{passaro2017informed}, we asked models to identify only coarse-grained entities: generic ones, such as persons (\textsc{PER}), locations (\textsc{LOC}), and organizations (\textsc{ORG});  and related to the administrative domain: law references (\textsc{LAW}), administrative acts (\textsc{ACT}), and PA organizations (\textsc{OPA}).

We prompted the models by feeding them with a sentence and an entity occurring in it, asking to predict the entity type in place of a masking token. The resulting template is: \texttt{\xml{text}. In questa frase, \xml{entity} è un esempio di \xml{mask}.}\footnote{In English: \texttt{\xml{text}. In this sentence, \xml{entity} is an example of \xml{mask}.}}

As anticipated, we verbalized the entities in three ways. In the first experiment, we provided an Italian translation of the entity or a single word representing the entity class. In the second experiment, we expanded most of the label words by including synonyms and other terms related to the various classes (See verbalizers for entity typing in Table \ref{et-verbalizer}). Finally, we used the knowledgeable verbalizer, by exploiting an external knowledge base automatically created as described in Section \ref{sec:expsetverb}.

\subsubsection{Document Classification} \label{sec:task-docclass}

For the recognition of the topics in PA and legal documents, we designed the following template to model the document classification task as a masked language modeling problem: 
\texttt{\xml{text}.Questo documento parla di \xml{mask}.}\footnote{In English: \texttt{\xml{text}. This document is about \xml{mask}}.}

Thus, LMs are challenged to infer the topic of the document by predicting the most appropriate label word to represent the masked token in the prompt, following the document text. Since the ATTO corpus contains only short documents of a maximum of 600 tokens, by setting the tokenizer's truncation at 512 tokens,\footnote{512 is the maximum number of tokens that these Transformers models can receive as input.} we were able to feed the models the entire document in almost all cases. To classify Italian civil judgements, we addressed the input length limit by feeding the models only with the factual background section of the judgements, which  provides a general overview of the topic.

\subsection{Evaluation metrics}\label{sec:evalmetrics}
We evaluated the performance of the models with common classification metrics, such as \textbf{Precision}, \textbf{Recall}, and \textbf{F1-Score}.

\subsection{Pseudo Log-Likelihood}

We employed Pseudo Log-Likelihood scores (PPLs) to evaluate the linguistic competence of models on domain texts with an intrinsic metric, and to determine whether this competence influenced their performance in domain-specific tasks modeled via prompting. Working with domain data necessitates understanding how well models can comprehend texts specific to that domain, especially when tasks are carried out without fine-tuning, so models can solely rely on the linguistic competence gained during pre-training to solve them. By incorporating PLLs computation into our analysis, we can gain insight into the capacity of our three models to represent and process domain specific terminology and linguistic structures.

The two specialized models, BureauBERTo and Ita-legal-BERT, are not only based on different encoder architecture (RoBERTa vs. BERT, respectively), but also diverge in terms of approaches used for their domain-adaptation: BureauBERTo vocabulary was expanded with domain-related tokens before additional pre-training, while no vocabulary expansion was performed for Ita-legal-BERT. Additionally, UmBERTo represents an instance of Italian models trained once and directly on a composite corpus comprising, among varoius genres, administrative and legal texts, which contributed to its effectiveness in tasks within these domains. Accounting for variability in models overall capacity to handle administrative and legal data, also in relation to the way they acquire their generic vs. domain-specific linguistic competence, can provide deeper insights into their behaviour when challenged on domain text classification tasks.

Thus, we aim to measure the probability given by the models to the tokens extracted from a sample of the considered datasets, or in other words, how much the models are surprised by analyzing the in-domain texts of our legal and PA datasets. Since we are dealing with encoder models, PLL is a valid alternative to Perplexity, which applies only to auto-regressive models. PLL \cite{salazar2020masked} is computed by summing the conditional log probabilities $\log P_{\text{MLM}}(w_t \mid W_{\setminus t})$ of each token $t$ in the sentence $W$. These are induced in BERT-like models by replacing $w_t$ with [MASK]. Let $\Theta$ indicate the model's parameter, then the PLL function is defined as:

\[
\text{PLL}(W) := \sum_{t=1}^{|W|} \log P_{\text{MLM}}(w_t \mid W_{\setminus t}; \Theta)
\]

To compute PLLs, we collected eight documents per class from the ATTO corpus and the judgements dataset, and 90 and 135 sentences per class from the InformedPA and the legal entity typing dataset, respectively (see Section \ref{sec:datasets}). 
The dataset size was chosen based on the number of elements in the least numerous class of the dataset and was stratified across all other classes. Additionally, we computed PLLs on a collection of 482 sentences from various genres, corresponding to the test set of the Italian Stanford Dependency Treebank (ISDT) \cite{bosco2014evalita}, that we used as baseline to identify any significant difficulties in modeling domain-specific texts.

Since PLLs are comparable under the same vocabulary dimension, which varied among our models, we normalized the scores using sentence length (i.e., the number of tokens) to mitigate this effect.

\section{Results and discussion}   \label{sec:results}

\subsection{Entity typing}

\subsubsection{PA domain}
In this paragraph, we analyze the results of the entity typing task when applied to the PA domain. Table \ref{tab:et-f1-amm} shows the Macro Average F1-score obtained by the three examined models in each setting, using the three verbalizers described earlier. In Table \ref{tab:et-f1-amm-classes} the F1-scores for each class are reported.

\textbf{No-calibration.} By employing a base verbalizer, where a single label is associated with each class, UmBERTo almost doubled the results obtained by BureauBERTo (0.36 vs. 0.20), whereas Ita-Legal-BERT obtained very low results (0.08). Surprisingly, for a domain entity like \textsc{ACT}, BureauBERTo missed all the entities, whereas UmBERTo obtained a low but higher score (0.14). For the \textsc{LAW} entity, UmBERTo outperforms both Ita-Legal-BERT and BureauBERTo, as well. We may suppose that this is because UmBERTo was trained on Common-Crawl, which also contains legal and administrative texts in its Italian section. Very high results are obtained by UmBERTo for \textsc{PER} entities, reaching 0.83 in our zero-shot scenario. Even though on \textsc{OPA} Ita-Legal-BERT overpasses BureauBERTo and UmBERTo (0.20 vs 0.05 and 0.19), the three models obtain very low results for this class and for \textsc{LOC}, and \textsc{ORG}. These two latter classes are very similar to each other: ORG refers to organizations in general, comprising firms and associations, whereas \textsc{OPA} can be considered as a subclass of \textsc{ORG}, and refers to organizations within the Public Administration, such as municipal departments. Such overlapping may impact classification.

Then, we added to the prompt  highly distinctive words for each class and integrated them into a manual verbalizer. In this case, we notice a better ability of both BureauBERTo and Ita-Legal-BERT to recognize domain-specific entities such as  \textsc{ACT}, \textsc{LAW}, and  \textsc{OPA}. However, despite the general improvement in recognizing such classes, we noticed that they perform worse than UmBERTo for traditional entities. Nonetheless, whereas Ita-Legal-BERT still obtained very low scores (F1 Marco AVG 0.15), BureauBERTo (0.52) outperforms the general-purpose model, UmBERTo (0.37), by exhibiting an improvement more than twofold (0.52 vs 0.20). This significant boost in performance suggests that the domain-adapted model is likely to be more attuned and proficient in leveraging domain-specific terminology.
 
Nevertheless, it is important to acknowledge that domain-specific terms may wield less influence over generic entities such as \textsc{PER}. With the manual verbalizer, UmBERTo fails to recognize any \textsc{PER} entity. We observed that the model identifies almost all the people's names as \textsc{ORG} entities. Thus, we carried out an ablation study by considering only the in-domain model BureauBERTo and the general-purpose model UmBERTo. The ablation study consisted in deleting the in-domain terms added for the \textsc{PER} entity class, i.e. \textit{generalità - ``particulars''} and \textit{nominativo - ``name''}. The results are summarized in Table \ref{tab:as-results}, Appendix \ref{ap:ablation}. They show that the performance of UmBERTo increases not only for the \textsc{PER} entities but that the ablation improves the F1-score of the \textsc{ORG} class as well. Whereas UmBERTo reaches the highest performances for overall F1 Micro Avg, the adapted model still obtained higher results on the in-domain entity classes: \textsc{ACT}, \textsc{LAW}, and \textsc{OPA},
further solidifying the advantages of domain-adapted models in specialized contexts. 

Given the high variability of results obtained by manually chosen terms composing the verbalizer, we decided to adopt knowledgeable verbalizers. We built a verbalizer for each model, according to the models' vocabulary (see Section \ref{sec:expsetverb}). However, in this case, the results dropped considerably: BureauBERto and UmBERTo reached even lower scores (0.12 and 0.28, respectively) than those obtained by using the base verbalizer. The lower results were still obtained for the \textsc{ORG} class, and in this case, also for the \textsc{LAW} class. In order to mitigate the effect of the terms related to a certain class over the identification of terms of other classes, we resort to the model's calibration (see Section \ref{sec:expsetcal}).

\textbf{Contextual calibration.} By adopting the CC, all models increase their scores, with two exceptions: BureauBERTo with the manual verbalizer decreases its F1-score from 0.52 to 0.46, and UmBERto does not increase its result with the knowledgeable verbalizer (0.28). BureauBERTo reaches the higher score in this setting with the knowledgeable verbalizer (0.58), but it reaches a higher score than the other models also with the base verbalizer, which means that in the absence of any other resource (knowledge base or in-domain text for calibration), the in-domain model is the most effective. Finally, Ita-Legal-BERT  obtains lower scores than the other two models with the base and the manual verbalizer, but exploits the information contained in the knowledgeable verbalizer better than UmBERTo (F1-score MacAVG. 0.52 vs. 0.28)).

In this setting, BureauBERTo obtains the best results for \textsc{ACT} and \textsc{LAW}. More specifically, the best results for \textsc{ACT} are obtained with the knowledgable verbalizer, whereas the \textsc{LAW} entities are better captured with the base verbalizer. Probably, references to other administrative acts have a more variable structure since they might rely more on each local administration so that the external knowledge base is more effective than with references to laws, which might have a more fixed structure. UmBERTo is still the best model on general entities and obtained the higher score also for the \textsc{OPA} class, by using the knowledgeable verbalizer.

\textbf{Batch calibration.} Surprisingly, another scenario unfolds when applying the BC. Even though Ita-Legal-BERT still obtains the lower scores, this time with all the verbalizers, it is UmBERTo to benefit the most from employing dataset instances for the calibration, by obtaining the overall highest scores. It reaches 0.62 F1-score MacroAvg both with the base and the manual verbalizer and still surpasses the other two models by using the knowledgeable verbalizer. 

In this scenario, UmBERTo reaches the highest score for \textsc{LOC} (0.85), \textsc{PER} (0.84), and \textsc{ORG} (0.43). It is worth noticing that UmBERTo doubled its performance with the knowledge verbalizer in the BC scenario.

\begin{table*}
\centering \small
\caption{\label{tab:et-f1-amm} F1-score obtained on Entity Typing in the administrative domain. Bold indicates the best result for each type of calibration, the best result for each model is underlined, and the best overall results are in italic.}
\begin{tabular}{@{}llccc@{}}
\toprule
\textbf{Calibration} & \textbf{Verbalizer} & \textbf{Ita-Legal-BERT} & \textbf{BureauBERTo} & \textbf{UmBERTo} \\
\midrule
\multirow{3}{*}{\textbf{No}} & Base & 0.08 & 0.20 & \textbf{0.36}\\
& Manual & 0.15 & \textbf{0.52} & 0.37\\
& Knowledgeable & 0.11 & 0.12 & \textbf{0.28}\\
\midrule
\multirow{3}{*}{\textbf{Contextual}} & Base & 0.17 & \textbf{0.54} & 0.50\\
& Manual & 0.22 & 0.46 & \textbf{0.50}\\
& Knowledgeable & 0.52 & \textbf{0.58} & 0.28\\
\midrule
\multirow{3}{*}{\textbf{Batch}} & Base & 0.26 & \underline{0.60} & \underline{\textit{\textbf{0.62}}}\\
& Manual & 0.32 & 0.59 & \underline{\textit{\textbf{0.62}}}\\
& Knowledgeable & \underline{0.42} & 0.55 & \textbf{0.56}\\
\bottomrule
\end{tabular}
\end{table*}

\begin{table*}
\centering \small
\caption{\label{tab:et-f1-amm-classes} F1-score obtained for each entity class on Entity Typing in the administrative domain. Bold indicates the best result for each type of calibration, the best result for each model is underlined, and the best overall results are in italic.}
\begin{tabular}{@{}lccc|ccc|ccc@{}}
\toprule
Label &  \multicolumn{3}{c}{\textbf{Base Verb.}} &  \multicolumn{3}{c}{\textbf{Manual Verb.}}  & \multicolumn{3}{c}{\textbf{Knowl. Verb.}} \\
 \midrule
  & UmB  & BB & ILB & UmB & BB & ILB & UmB & BB & ILB \\
\midrule
 &  \multicolumn{9}{c}{No-Calibration} \\
\midrule
\textsc{LOC} & 0.09 & 0.03 & 0.00 & \textbf{0.56} &  0.51  & 0.00  & 0.38 & 0.24 & 0.01\\
\textsc{ORG} & 0.24 & 0.15 & 0.00 & 0.15 & 0.21 & \textbf{0.27} & 0.09 & 0.04 & 0.00\\
\textsc{PER} & \textbf{0.83} & 0.63 & 0.23 & 0.00 & 0.50 & 0.00 & 0.47 & 0.21 & 0.38\\
\textsc{ACT} & 0.14 & 0.00 & 0.00 & 0.41 & \textbf{0.46} & 0.20  & 0.16 & 0.02 & 0.01\\
\textsc{LAW} & 0.52 & 0.36 & 0.07 & 0.67 & \textbf{0.74} & 0.30  & 0.02 & 0.00 & 0.00\\
\textsc{OPA} & 0.19 & 0.05 & 0.20 & 0.41 & \textbf{0.69}  & 0.16  & 0.56 & 0.23 & 0.27\\

\midrule
 &  \multicolumn{9}{c}{Contextual Calibration} \\
\midrule
LOC & 0.45 & 0.59 & 0.01 & \textbf{0.75} & 0.59 & 0.08 & 0.56 & 0.62 & 0.17 \\
ORG & 0.34 & 0.33 & 0.19 & \underline{\textbf{\textit{0.43}}} & 0.19 & 0.20 & 0.40 & \underline{0.35} & 0.08 \\
PER & 0.70 & \underline{0.71} & 0.30 & 0.31 & 0.22 & 0.04 & \textbf{0.74} & 0.69 & \underline{0.49} \\
ACT & 0.51 & 0.45 & 0.00 & \underline{0.54} & 0.58 & 0.27 & 0.46 & \underline{\textit{\textbf{0.60}}} & 0.40 \\
LAW & 0.49 & \underline{\textbf{\textit{0.73}}} & 0.41 & \underline{0.66} & 0.61 & 0.46 & 0.23 & 0.64 & 0.21 \\
OPA & 0.46 & 0.42 & 0.19 & 0.32 & 0.55 & 0.26 & \underline{\textit{\textbf{0.74}}} & 0.60 & 0.33 \\

\midrule
 & \multicolumn{9}{c}{Batch Calibration} \\
\midrule
LOC & 0.77 & \underline{0.75} & 0.18 & \underline{\textbf{\textit{0.85}}} & 0.72 & 0.17 & 0.80 & 0.71 & \underline{0.36} \\
ORG & \underline{\textbf{\textit{0.43}}} & 0.31 & 0.10 & 0.41 & 0.24 & \underline{0.23} & 0.22 & 0.19 & \underline{0.20} \\
PER & \underline{\textbf{\textit{0.84}}} & \underline{0.71} & 0.33 & 0.80 & 0.67 & 0.28 & 0.77 & 0.62 & 0.44 \\
ACT & 0.53 & \textbf{0.59} & 0.02 & 0.45 & 0.56 & 0.27 & 0.50 & 0.57 & \underline{0.54} \\
LAW & 0.54 & 0.64 & 0.56 & 0.56 & \textbf{0.70} & \underline{0.61} & 0.41 & 0.62 & 0.44 \\
OPA & 0.62 & 0.61 & 0.39 & 0.67 & \underline{0.65} & 0.33 & \textbf{0.69} & 0.56 & \underline{0.57} \\

 \bottomrule
\end{tabular}
\end{table*}

\subsubsection{Legal domain}

In this paragraph, we analyze the results of entity typing modeled as a prompting task within the legal domain. Table \ref{tab:et-f1-legal} presents the Macro Average F1-score achieved by the three models under examination across different settings, utilizing the three verbalizers previously described. Table \ref{tab:et-f1-legal-classes} details the F1-scores for each class.

As for the PA domain, also with legal documents we observed that the models manage to benefit from the knowledgeable verbalizer only when calibration is also applied. In fact, the performance of the three models worsened with the knowledgeable verbalizer without calibration, whereas the base verbalizer allowed for reaching only 0.29 F1-score MacAvg by UmBERTo, followed by Ita-Legal-BERT and then by BureauBERTo. Surprisingly, similarly to the PA domain, it is the general-purpose model to obtain the best score for this setting only by exploiting a single label per class as in the base verbalizer.

With the CC, as BureauBERTo for the PA domain, Ita-Legal-BERT reaches the best score when using only the base verbalizer, highlighting the relevance of a domain-specialized model when lacking other in-domain resources (e.g., corpora or knowledge bases). However, in this case, Ita-Legal-BERT reaches a very low score (0.22), and it is surpassed by both BureauBERTo and UmBERTo when using a knowledgeable verbalizer, reaching 0.33 F1-score. Nonetheless, by using only the base verbalizer, Ita-Legal-BERT reaches the best overall score for the \textsc{LAW} entities.

The best scores for the other entities are reached in the third setting, the BC. The names of judges and the names of lawyers are better identified by the domain-specialized models: Ita-Legal-BERT, by using only the base verbalizer, and BureauBERTo, with the knowledgeable verbalizer, respectively. References to judgments are best identified by UmBERTo with the knowledgeable verbalizer, even though with a still lower F1-score (0.30). To conclude, we can observe that the models not specialized in the legal domain manage to obtain good results with the combination of the external knowledge base and BC. Whereas Ita-Legal-BERT reaches the overall highest score by only using the base verbalizer, obtaining 0.41 on par with BureauBERTo.

\begin{table*}
\centering \small
\caption{\label{tab:et-f1-legal} F1-score obtained on Entity Typing in the legal domain. Bold indicates the best result for each type of calibration, the best result for each model is underlined, and the best overall results are in italic.}
\begin{tabular}{@{}llccc@{}}
\toprule
\textbf{Calibration} & \textbf{Verbalizer} & \textbf{Ita-Legal-BERT} & \textbf{BureauBERTo} & \textbf{UmBERTo} \\
\midrule
\multirow{2}{*}{\textbf{No}} & Base & 0.20 & 0.15 & \textbf{0.29}\\
& Knowledgeable & \textbf{0.08} & 0.05 & 0.06\\
\midrule
\multirow{2}{*}{\textbf{Contextual}} & Base & \textbf{0.22} & 0.17 & 0.20 \\
& Knowledgeable & 0.24 & \textbf{0.33} & \textbf{0.33}\\
\midrule
\multirow{2}{*}{\textbf{Batch}} & Base & \underline{\textit{\textbf{0.41}}} & \textbf{0.37} & 0.22\\
& Knowledgeable & 0.37 & \underline{\textit{\textbf{0.41}}} & \underline{\textbf{0.40}}\\

\bottomrule
\end{tabular}
\end{table*}

\begin{table*}
\centering \small
\caption{\label{tab:et-f1-legal-classes} F1-score obtained for each entity class on Entity Typing in the legal domain. Bold indicates the best result for each type of calibration, the best result for each model is underlined, and the best overall results are in italic.}
\begin{tabular}{@{}lccc|ccc@{}}
\toprule
\textbf{Label} &  \multicolumn{3}{c}{\textbf{Base Verb.}}  & \multicolumn{3}{c}{\textbf{Knowledgeable Verb.}} \\
 \midrule
  & UmBERTo  & BureauBERTo & Ita-LB & UmBERTo & BureauBERTo & Ita-LB \\
\midrule
 &  \multicolumn{6}{c}{Contextual Calibration} \\
\midrule
\textsc{JDG} & 0.20 & 0.21 & 0.00 & 0.50 & \textbf{0.58} & 0.00 \\
\textsc{LAW} & 0.15 & 0.06 & \underline{\textbf{\textit{0.43}}} & \underline{0.28} & 0.19 & \underline{0.38} \\
\textsc{LWY} & 0.31 & 0.35 & 0.24 & 0.34 & \textbf{0.42} & 0.41 \\
\textsc{RUL} & 0.11 & 0.04 & \textbf{0.20} & 0.18 & 0.13 & 0.18 \\
\midrule
 & \multicolumn{6}{c}{Batch Calibration} \\
\midrule
\textsc{JDG} & 0.20 & 0.43 & 0.41 & \underline{0.55} & \underline{\textbf{\textit{0.67}}} & \underline{0.42} \\
\textsc{LAW} & 0.18 & 0.23 & \textbf{0.36} & 0.27 & \underline{0.25} & \textbf{0.36} \\
\textsc{LWY} & 0.31 & \underline{0.57} & \underline{\textbf{\textit{0.60}}} & \underline{0.50} & 0.54 & 0.47 \\
\textsc{RUL} & 0.18 & \underline{0.24} & \underline{0.24} & \underline{\textbf{\textit{0.30}}} & 0.19 & 0.23 \\

 \bottomrule
\end{tabular}
\end{table*}

 \subsection{Document classification}

\subsubsection{PA domain}

Regarding the prompt document classification experiments in the PA domain, the results summarized in Table \ref{tab-DC-PA} reveal a trend similar to that observed in entity typing.

\textbf{No-calibration.} Using the base verbalizer, where only one or few word labels represent a topic class, the domain-specialized model BureauBERTo, and the legal-specialized model Ita-Legal-Bert achieved rather low Macro Avg F1-scores (0.06 vs. 0.07), when tested without any calibration. In contrast, the generic model UmBERTo obtained a slightly higher F1 score of 0.16, outperforming the other two models in almost all classes, except for \textsc{Cultura, Turismo e Sport - `Culture, tourism, and sports'}, \textsc{Demografico - `Demographics'}, and \textsc{Personale - `Personnel'} (see Table \ref{tab:class-f1-amm-class}). 

Conversely, when prompted using the manual and the knowledgeable verbalizers, which included a set of salient PA-related terms 
to depict the document topics at a finer-grained level, we observed a significant improvement in the overall performance of all models. The benefits of a custom-made or automatically constructed set of domain-related terms were particularly evident for the specialized model BureauBERTo, which achieved the highest Macro Avg F1-scores in both settings (0.35 and 0.36, respectively). It appears that the model adapted to the PA domain may possess heightened sensitivity, enabling it to effectively capitalize on the contextual cues provided by domain-specific terms. However, with the manual verbalizer, we observed that for some classes sharing a common domain lexicon, such as \textsc{Pubblica Istruzione - `Public Education'} and \textsc{Cultura, Turismo e Sport - `Culture, Tourism, and Sports'}, or \textsc{Servizi finanziari - `Financial services'} and \textsc{Bandi e Contratti - `Tenders and Contracts'} the domain models' classification could have been biased
in favor of one of the two classes.

These findings highlight the necessity of addressing models' biases when performing prompt-based classification tasks, especially with domain-specialized models like BureauBERTo, whose vocabulary distribution has been altered due to vocabulary expansion and additional pre-training on domain-specific corpora.

\textbf{Contextual calibration.} To mitigate such biases we repeated the experiments with the three verbalizers, adding contextual calibration. As shown in Table \ref{tab-DC-PA}, calibrating the models on content-free input helped enhance the models' performance. In this case, the F1-scores obtained with the base verbalizer by BureauBERTo (0.20), UmBERTo (0.14), and Ita-Legal-BERT (0.12) improved compared to those achieved without calibration. Notably, contrary to previous findings, BureauBERTo emerged as the best-performing model with the base verbalizer, while the generic UmBERTo achieved the highest overall result (0.35) with the manual verbalizer, followed by Ita-Legal-BERT (0.31) and BureauBERTo (0.20). We speculate that, once the inherent models propensity to favor classes with frequent labels is mitigated by CC, manually crafting the verbalizer for specialized models may not always be the most favorable choice. During additional pre-training on domain texts, BureauBERTo encountered words like \textit{contracts, services, goods, residents,} etc., multiple times and in relation to different topics. Consequently, these words may not be as closely associated with a specific topic as they appear to a human annotator, potentially disrupting the model behavior.

Indeed, a different scenario emerges when using the knowledgeable verbalizer, where we leveraged the model itself to identify the most appropriate word label candidates. In this case, BureauBERTo, achieved the highest performance (with a Macro Avg F1-score of 0.32), followed by Ita-legal-BERT, which typically performs worse than UmBERTo, but here obtained a better result (0.28 of Macro Avg F1-score). Finally, UmBERTo, was the worst-performing model in this setting (with a Macro Avg F1-score of 0.25).

\textbf{Batch calibration.} Compared to CC, BC brought a higher gain in almost all models  with all three types of verbalizers. Unfortunately, this approach is only practicable when a sufficient amount of data is available to calibrate the models, which is quite uncommon for specific domains. For the PA domain, we used a sample of 200 unlabelled documents stratified across the thirteen classes in our dataset.

With the base verbalizer, we observed generally enhanced performance compared to the BC and the no-calibration settings. In this context, the best-performing model was UmBERTo (0.32 Macro Avg F1-score), which also achieved the highest performance with the manual verbalizer (0.45). The only scenario in which the domain-specialized model BureauBERTo surpassed UmBERTo was with the knowledgeable verbalizer (0.40 vs. 0.35, respectively). Furthermore, calibrating the models with a context-based approach, i.e., on instances of the dataset, resulted in higher performance with the manual verbalizer than with the knowledgeable verbalizers for all three models. 
This suggests that when there's an opportunity to calibrate models on real domain data, leveraging human expertise to construct the verbalizer can yield better overall results, especially when a domain-specialized model is not available. 
However, the fact that the generic model UmBERTo performs better than BureauBERTo with the base and manual verbalizer, but not with the knowledgeable one, supports the hypothesis that when working with specialized models, especially when no additional data are available for calibration, it is better to select the appropriate labels for the task by leveraging the model itself.

\begin{table*}
\centering \small
\caption{\label{tab-DC-PA} F1-score obtained on Document Classification in the administrative domain. Bold indicates the best result for each type of calibration, the best result for each model is underlined, and the best overall results are in italic.}
\begin{tabular}{@{}llccc@{}}
\toprule
\textbf{Calibration} & \textbf{Verbalizer} & \textbf{Ita-Legal-BERT} & \textbf{BureauBERTo}  & \textbf{UmBERTo}  \\
\midrule
\multirow{3}{*}{\textbf{No}} & Base & 0.07 & 0.06 &  \textbf{0.16}  \\
& Manual & 0.19  & \textbf{0.35} & 0.32  \\
& Knowledgeable & 0.18  & \textbf{0.36} & 0.26  \\
\midrule
\multirow{3}{*}{\textbf{Contextual}} & Base & 0.12 & \textbf{0.21} & 0.13 \\
& Manual & 0.31 & 0.20 & \textbf{0.35} \\
& Knowledgeable & 0.28 & \textbf{0.32} & 0.25 \\
\midrule
\multirow{3}{*}{\textbf{Batch}} & Base & 0.16 & 0.31 & \textbf{0.32}\\
& Manual & \underline{0.33} & \underline{0.42} & \underline{\textit{\textbf{0.45}}} \\
& Knowledgeable & 0.22 & \textbf{0.40} & 0.35 \\
\bottomrule
\end{tabular}
\end{table*}

\begin{table*}
\centering \small
\caption{\label{tab:class-f1-amm-class} F1-score obtained for each entity class on Document Classification in the administrative domain. Bold indicates the best result for each type of calibration, the best result for each model is underlined, and the best overall results are in italic.\textsc{C-A-E} refers to \textsc{Commercio-Attività-Economiche}, whereas \textsc{C-T-S} stands for \textsc{Cultura-Turismo-Sport}.}
\begin{tabular}{@{}lccc|ccc|ccc@{}}
\toprule
\textbf{Label} &  \multicolumn{3}{c}{\textbf{Base Verb.}} &  \multicolumn{3}{c}{\textbf{Manual Verb.}}  & \multicolumn{3}{c}{\textbf{Knowl. Verb.}} \\
 \midrule
  & UmB  & BB & ILB & UmB & BB & ILB & UmB & BB & ILB \\
\midrule
 &  \multicolumn{9}{c}{Contextual Calibration} \\
 \midrule
\textsc{C-A-E} & 0.0 & 0.18 & 0.00 & \underline{\textit{\textbf{0.30}}} & \underline{0.25} & 0.00 & 0.00 & 0.11 & 0.00 \\
\textsc{Demografico} & 0.06 & 0.15 & 0.02 & 0.52 & 0.17 & 0.44 & \underline{\textit{\textbf{0.76}}} & 0.40 & \underline{0.62} \\
\textsc{Personale} & 0.16 & 0.33 & 0.18 & \textbf{0.42} & 0.37 & 0.26 & 0.01 & 0.13 & 0.11 \\
\textsc{Avvocatura} & 0.10 & 0.07 & 0.04 & \underline{0.13} & 0.04 & 0.22 & 0.08 & \textbf{0.36} & 0.02 \\
\textsc{Servizi-informativi} & 0.00 & 0.4 & 0.03 & \underline{\textbf{\textit{0.19}}} & 0.00 & 0.01 & 0.00 & 0.00 & 0.03 \\
\textsc{C-T-S} & 0.23 & 0.28 & 0.02 & \textbf{0.30} & 0.16 & 0.20 & 0.22 & 0.25 & 0.23 \\
\textsc{Ambiente} & 0.01 & 0.03 & 0.21 & \textbf{0.42} & 0.01 & \underline{0.33} & 0.17 & 0.17 & 0.20 \\
\textsc{Sociale} & 0.01 & 0.15 & 0.01 & 0.04 & 0.00 & 0.14 & \underline{\textbf{\textit{0.29}}} & 0.02 & \underline{0.20} \\
\textsc{Bandi-contratti} & 0.14 & 0.09 & 0.00 & \textbf{0.15} & 0.13 & 0.09 & 0.03 & 0.07 & 0.02 \\
\textsc{Servizio-finanziario} & 0.46 & 0.00 & 0.62 & \textbf{0.72} & 0.11 & 0.54 & 0.70 & 0.58 & 0.58 \\
\textsc{Urbanistica} & 0.40 & 0.62 & 0.01 & \underline{0.81} & 0.61 & 0.77 & 0.51 & \textbf{0.83} & 0.79 \\
\textsc{Pubblica-istruzione} & 0.00 & 0.55 & 0.22 & 0.00 & 0.28 & \underline{\textbf{\textit{0.60}}} & 0.07 & \underline{0.59} & 0.54 \\
\textsc{Edilizia} & 0.17 & 0.17 & 0.24 & 0.60 & 0.41 & \underline{0.37} & 0.46 & \textbf{0.61} & 0.35 \\

\midrule
 & \multicolumn{9}{c}{Batch Calibration} \\
 \midrule
\textsc{C-A-E} & 0.18 & 0.12 & \underline{0.02} & \textbf{0.20} & 0.15 & 0.06 & 0.00 & 0.00 & 0.01 \\
\textsc{Demografico} & 0.26 & 0.10 & 0.23 & \textbf{0.73} & 0.55 & 0.58 & 0.60 & \underline{0.67} & 0.53 \\
\textsc{Personale} & 0.36 & 0.58 & 0.27 & \underline{\textbf{\textit{0.60}}} & \underline{\textbf{\textit{0.60}}}  & \underline{0.36} & 0.14 & 0.48 & 0.17 \\
\textsc{Avvocatura} & 0.09 & 0.12 & 0.04 & 0.11 & \underline{\textbf{\textit{0.42}}} & \underline{0.37} & 0.09 & 0.33 & 0.27 \\
\textsc{Servizi-informativi} & \textbf{0.12} & 0.04 & \underline{0.07} & 0.07 & 0.02 & 0.02 & 0.10 & \underline{0.08} & 0.02 \\
\textsc{C-T-S} & 0.26 & 0.24 & 0.06 & \underline{\textbf{\textit{0.34}}} & 0.26 & \underline{0.26} & 0.23 & 0.18 & 0.18 \\
\textsc{Ambiente} & 0.45 & 0.26 & 0.15 & 0.59 & \underline{\textbf{\textit{0.61}}} & 0.21 & \underline{\textbf{\textit{0.61}}} & 0.58 & 0.27 \\
\textsc{Sociale} & 0.12 & 0.25 & 0.03 & 0.18 & \underline{\textbf{0.26}} & 0.03 & 0.05 & 0.13 & 0.13 \\
\textsc{Bandi-contratti} & \underline{0.16} & 0.11 & 0.00 & 0.17 & \underline{\textbf{\textit{0.22}}} & \underline{0.12} & 0.14 & 0.04 & 0.07 \\
\textsc{Servizio-finanziario} & 0.79 & 0.67 & 0.64 & \underline{\textbf{\textit{0.84}}} & \underline{0.77} & \underline{0.69} & 0.76 & 0.74 & 0.05 \\
\textsc{Urbanistica} & 0.68 & 0.53 & 0.26 & 0.79 & \underline{\textbf{\textit{0.87}}} & \underline{0.80} & 0.60 & 0.77 & 0.76 \\
\textsc{Pubblica-istruzione} & 0.30 & 0.46 & 0.07 & \underline{\textbf{0.55}} & 0.22 & 0.52 & 0.54 & 0.51 & 0.10 \\
\textsc{Edilizia} & 0.43 & 0.50 & 0.20 & 0.66 & 0.53 & 0.30 & \underline{\textbf{\textit{0.67}}} & \underline{\textbf{\textit{0.67}}} & 0.36 \\

 \bottomrule
\end{tabular}
\end{table*}

\subsubsection{Legal Domain}
Regarding the prompt-document classification in the legal domain, we only tested models using the base and the knowledgeable verbalizers, as we lacked the domain expertise necessary to select the most suitable labels to best depict the different topics in the civil judgments dataset for the manual verbalizer.

As shown in Table \ref{tab:class-f1-legal}, with the base verbalizer, the best-performing model was consistently BureauBERTo, followed by UmBERTo and Ita-legal-BERT, across all three settings: no-calibration, CC and BC. Despite being the domain-specialized model, Ita-legal-BERT struggles to reach the accuracy levels of the other two models when tasked via prompting. This may be because Ita-legal-BERT is based on the BERT architecture, while the other two models can rely on the robustly optimized architecture of RoBERTa \cite{liu2019roberta}. The advantage of BureauBERTo over UmBERTo in classifying legal texts may stem from its additional pre-training on administrative texts, which in the Italian language share some common linguistic features with legal language.

On the other hand, with the knowledgeable verbalizer, the domain-specialized model outperforms the other two in the CC setting, namely when the word labels are selected by the model and calibration is performed on content-free input strings. In contrast, when calibrating on data extracted from the dataset (BC), the best performing model was UmBERTo. The class-wise comparison reported in Table \ref{tab:class-f1-legal-class} shows that BC actually improved UmBERTo performance by enhancing its ability to recognize classes such as \textsc{edilizia e urbanistica}-\textsc{'constructions and urban planning'}, \textsc{servizi demografici}-\textsc{'demographics'} or \textsc{personale}-\textsc{'personnel'} which the model completely fails to categorize in the CC setting.
Similarly, for BureauBERTo and Ita-Legal-BERT, we observed a better ability to recognize most of the classification classes in the BC setting compared to their perfomance with CC. The least recognizable class for all three models was \textsc{amministrazione e segreteria generale}-\textsc{'administration and general secretariat'}, likely because the texts belonging to this class cover a wide range of diverse and generic topics.

This confirms that BC calibration is a powerful method to stabilize and improve model performance, enabling the effective use of even generic models for domain specific task.

\begin{table*}
\centering \small
\caption{\label{tab:class-f1-legal} F1-score obtained on Document Classification in the legal domain. Bold indicates the best result for each type of calibration, the best result for each model is underlined, and the best overall results are in italic.}
\begin{tabular}{@{}llcc|cc|cc@{}}
\toprule
\textbf{Calibration} & \textbf{Verbal.} & \multicolumn{2}{c}{\textbf{Ita-Legal-BERT}} & \multicolumn{2}{c}{\textbf{BureauBERTo}} & \multicolumn{2}{c}{\textbf{UmBERTo}} \\
 \midrule
 &  & MacAVG & WeAvg & MacAVG & WeAvg & MacAVG & WeAvg \\
\midrule
\multirow{3}{*}{\textbf{No}} & Base  & 0.15 & 0.17 & \textbf{0.27} & \textbf{0.28} & 0.26 & 0.27 \\
 & Knowl. & 0.13 & 0.21 & \textbf{0.20} & \textbf{0.26} & 0.19 & 0.22\\
 \midrule
 \multirow{3}{*}{\textbf{Contextual}} & Base  & 0.15 & 0.19 &\textbf{0.30} & \textbf{0.37} & 0.22 & 0.25 \\
& Knowl. & \textbf{0.19} & \textbf{0.26} & 0.15 & 0.15 & 0.11 & 0.12\\
 \midrule
\multirow{3}{*}{\textbf{Batch}} & Base & 0.25 & 0.33 & \underline{\textit{\textbf{0.41}}} & \underline{\textbf{0.47}} & 0.39 & 0.46 \\
 & Knowl. & \underline{0.27} & \underline{0.34} & 0.39 & 0.46 & \underline{\textit{\textbf{0.41}}} & \underline{\textit{\textbf{0.50}}} \\
 \bottomrule
\end{tabular}
\end{table*}

\begin{table*}
\centering \small
\caption{\label{tab:class-f1-legal-class} F1-score obtained for each class on Document Classification in the legal domain. Bold indicates the best result for each type of calibration, the best result for each model is underlined, and the best overall results are in italic.}
\begin{tabular}{@{}lccc|ccc@{}}
\toprule
\textbf{Label} &  \multicolumn{3}{c}{\textbf{Base Verbalizer}}  & \multicolumn{3}{c}{\textbf{Knowledgeable Verbalizer}} \\
 \midrule
  & UmB  & BB & Ita-LB & UmB & BB & Ita-LB \\
\midrule
 &  \multicolumn{6}{c}{Contextual Calibration} \\
\midrule
\textsc{Polizia Locale} & 0.02 & 0.51 & 0.29 & 0.23 & 0.07 & \underline{\textbf{0.69}} \\
\textsc{Edilizia e Urbanistica} & 0.63 & \textbf{0.69} & 0.48 & 0.00 & 0.13 & 0.11 \\
\textsc{Attività Economiche} & 0.00 & 0.06 & 0.03 & 0.03 & 0.16 & \textbf{0.21} \\
\textsc{Appalti e Contratti} & \underline{\textbf{0.54}} & 0.43 & 0.00 & 0.23 & 0.42 & 0.34 \\
\textsc{Servizi Demografici} & 0.18 & 0.17 & 0.02 & 0.00 & \underline{\textbf{0.23}} & 0.03 \\
\textsc{Tributi Locali} & 0.26 & \textbf{0.39} & 0.17 & 0.36 & 0.01 & 0.01 \\
\textsc{Personale} & \textbf{0.38} & 0.31 & 0.17 & 0.00 & 0.20 & 0.16 \\
\textsc{Bilancio e Contabilità} & 0.01 & 0.08 & \textbf{0.14} & 0.10 & 0.04 & \textbf{0.14} \\
\textsc{Ammin. e Segreteria Gen.} & 0.00 & \underline{\textit{\textbf{0.08}}} & 0.06 & 0.00 & 0.05 & 0.04 \\

\midrule
 & \multicolumn{6}{c}{Batch Calibration} \\
\midrule
\textsc{Polizia Locale} & 0.66 & \underline{0.58} & 0.58 & \underline{\textit{\textbf{0.75}}} & 0.57 & 0.64 \\
\textsc{Edilizia e Urbanistica} & 0.66 & \underline{\textit{\textbf{0.71}}} & \underline{0.57} & \underline{0.69} & 0.68 & 0.31 \\
\textsc{Attività Economiche} & 0.11 & 0.20 & 0.07 & \underline{0.31} & \underline{\textit{\textbf{0.36}}} & \underline{0.22} \\
\textsc{Appalti e Contratti} & 0.64 & \underline{\textit{\textbf{0.70}}} & 0.19 & 0.52 & 0.52 & \underline{0.35} \\
\textsc{Servizi Demografici} & 0.19 & 0.15 & 0.12 & \underline{\textit{\textbf{0.30}}} & 0.19 & \underline{0.26} \\
\textsc{Tributi Locali} & \underline{0.49} & \underline{\textit{\textbf{0.54}}} & 0.16 & 0.33 & 0.47 & \underline{0.33} \\
\textsc{Personale} & 0.40 & \underline{0.42} & \underline{0.34} & \underline{\textit{\textbf{0.44}}} & \underline{0.42} & 0.05 \\
\textsc{Bilancio e Contabilità} & 0.32 & \underline{\textit{\textbf{0.29}}} & 0.22 & 0.27 & 0.26 & 0.19 \\
\textsc{Ammin. e Segreteria Gen.}  & \underline{0.07} & \underline{\textit{\textbf{0.08}}} & 0.02 & \underline{0.07} & 0.07 & \underline{0.07} \\

 \bottomrule
\end{tabular}
\end{table*}

\subsection{Pseudo-Log-Likelihood scores (PLLs)}

The PLLs obtained by BureauBERTo, Ita-Legal-Bert, and UmBERTo on the in-domain and general-purpose datasets are listed in Table \ref{tab:pllperformance_metrics}. Lower PLLs generally indicate better competence in modeling the texts. Our results showed that, across different corpora, BureauBERTo consistently achieves better (lower) PLLs compared to Ita-Legal-BERT and UmBERTo, with Ita-Legal-BERT generally having the highest PLLs. This aligns with Ita-Legal-BERT's lower performance on almost all classification tasks across several settings.

Although UmBERTo is a generic model, it handles better than Ita-Legal-BERT both PA and legal texts, likely due to its pre-training corpus, which included texts extracted from Italian government and municipalities web pages. This also explains why UmBERTo was the best generic model for handling administrative data \cite{auriemma2022evaluating} via fine-tuning and remains competitive on prompt-based tasks, particularly when calibrated on domain-specific content.

Surprisingly, the overall scores obtained by the three models on domain-specific datasets are lower (hence better) than those obtained on general-purpose texts. This finding may appear counterintuitive. However, assuming shared knowledge about Italian bureaucratic texts among all the models under our investigation, this outcome can be interpreted by considering the major variability in word distribution encountered in general-purpose texts compared to domain-specific documents.
In domain-specific language, such as legal or administrative texts, the vocabulary tends to be more formulaic and standardized, facilitating word prediction for the models. Conversely, general-purpose texts exhibit higher variability in topics and the words used to describe various situations and scenarios, posing a greater challenge for accurate word prediction. This increased variability may explain why the models perform worse on general-purpose texts, contrary to the expectation that domain-specific language would be more challenging to model. 
However, as evidenced by our results, performing a classification task on domain-specific data is still more challenging than simply predicting a masked word based on sentence context because assigning the correct label to a certain text requires deeper domain and task knowledge.

\begin{table}[h]
\centering \small
\caption{\label{tab:pllperformance_metrics} PLLs for Ita-Legal-BERT (ILB), UmBERTo (UmB), and BureauBERTo (BB) across in-doman and general purpose datasets.}
\centering
\begin{tabular}{@{}llccccc@{}}

\toprule
\textbf{Model} & \textbf{Measure} & \multicolumn{2}{c}{\textbf{PA}} & \multicolumn{2}{c}{\textbf{Legal}} & \textbf{General-purpose}\\
\midrule
 & & Doc Class & ET & Doc Class & ET & Treebank \\
\midrule
\multirow{2}{*}{ILB} & Mean & -1.86 & -1.75 & -1.13 & -1.84 & -3.98 \\
 & Std & 0.51 & 0.97 & 0.20 & 0.89 & 1.93 \\
\midrule
\multirow{2}{*}{UmB} & Mean & -1.26 & -2.10 & -0.97 & -2.05 & -2.87 \\
 & Std & 0.34 & 1.19 & 0.18 & 1.31 & 1.71 \\
\midrule
\multirow{2}{*}{BB} & Mean & -0.88 & -1.04 & -0.89 & -1.80 & -2.24 \\
 & Std & 0.30 & 0.83 & 0.17 & 1.20 & 1.38 \\
\bottomrule
\end{tabular}
\end{table}

\section{Conclusion and future work} \label{sec:conclusion}

In this paper, we propose a zero-shot prompt tuning classification approach for solving two tasks related to the Italian PA and legal domains: the classification of documents according to their topic and the recognition of the entity types occurring in administrative sentences.

We compared the perfomance of the PA-specialized model BureauBERTo with that of the legal-specialized Ita-Legal-BERT and of the domain-agnostic model UmBERTo across these tasks. 
Our findings indicate that, by performing calibration with both Contextual Calibration (CC) and Batch Calibration (BC) approaches, all models demonstrated enhanced performance in almost all settings. However, we observed an interplay between the type of calibration performed and the adopted verbalizer.

In the PA domain, when the task is performed by the specialized model with the knowledgeable verbalizer, calibrating the model on empty input strings (CC), yields satisfactory results, meaning that in this case the verbalizer can be built by leveraging self-supervision techniques, without human intervention. Similarly, for the classification of legal documents, the in-domain model Ita-Legal-BERT outperforms the other two only when the knowledgeable verbalizer is combined with the CC approach.

The knowledgeable verbalizer also allows BureauBERTo to achieve the highest results in the administrative text classification in both BC and CC settings, and even without calibration. Hence, another advantage of using a specialized model is the possibility of leveraging the model itself to construct an effective task verbalizer, thanks to the model ability to find appropriate in-domain word labels.

In contrast, with the manual verbalizer, which we only tested in the PA domain, the best-performing model was the generic UmBERTo in both BC and CC settings. This demonstrates that  domain-agnostic models can be effectively employed for solving domain-specific tasks when empowered by human domain expertise encompassed within the verbalizer and stabilized through calibration techniques. Adopting a manual verbalizer in combination with the CC approach can be particularly beneficial for specific domains  where neither specialized models nor domain data are available for calibration or to construct an external knowledge base from which to derive a knowledgeable verbalizer.

Among all, the BC approach enables the best overall performance for all three models across different tasks and domains. This supports previous findings indicating that relying on content-based strings instead of content-free input can help to better mitigate the model contextual biases in prompt-based tasks \cite{zhou2023batch}, particularly in zero-shot scenarios. However, this approach requires additional batches of class examples, which may not always be available in specific domains.

Furthermore, with the BC, we observed competitive performance in the entity typing tasks for both specialized models, BureauBERTo and Ita-Legal-BERT, in their respective domains, even when the tasks are performed using the base verbalizer. This suggests that for tasks like entity typing, where entities can be described using a small set of domain words, the domain-specialized models represent a valuable choice since they do not require additional human effort to construct the verbalizers.

In conclusion, our results underscore the possibility of effectively using encoder-only models for performing domain-related tasks modeled via prompting in a zero-shot setting, especially when stabilized through calibration techniques. Specialized encoders demonstrate competitive abilities in handling both generic and domain-specific data, as evidenced by their PPLs (see Table \ref{tab:pllperformance_metrics}), and obtain surprisingly high results also in supervised tasks, such as text classification. 
Notably, their acquired domain knowledge also enables these models to perform well in entity typing tasks without requiring additional effort to customize the verbalizers. These models are indeed accustomed to the distinctive way in which entity names and domain-specific terminology are employed in PA and legal texts.
This capability makes them considerably helpful for automatically finding finer-grained word labels to perform text classification without relying on annotated data. This opens up the possibility of employing smaller specialized models to automate the verbalizer construction, which might also be used to perform the task with larger models. 

As a future direction for our work, we aim to explore additional approaches to leverage specialized models for prompt-based tasks, fully exploiting their potential in contexts where annotated data are limited or unavailable. Additionally, we plan to test larger language models on these tasks to assess the impact of model size on domain-specific performance.

\backmatter

\bmhead{Acknowledgements}

This research has been supported by: The Project “ABI2LE (Ability to Learning)”, Regione Toscana (POR Fesr 2014-2020); PNRR - M4C2 - Investimento 1.3, Partenariato Esteso PE00000013 - ``FAIR - Future Artificial Intelligence Research" - Spoke 1 ``Human-centered AI", funded by the European Commission under the NextGeneration EU programme; TAILOR, funded by EU Horizon 2020 research and innovation programme under GA No 952215.

\bmhead{Author Contribution Statement}

Serena Auriemma: Writing – review \& editing, Writing – original draft, Visualization, Validation, Software, Methodology, Experiment Design, Data Analysis, Data curation, Conceptualization.
Martina Miliani:  Writing – review \& editing, Validation, Software, Methodology, Experiment Design, Data Analysis, Data curation, Conceptualization.
Mauro Madeddu: Validation, Software.
Alessandro Bondielli: Methodology, Validation, Software.
Alessandro Lenci: Writing – review \& editing, Methodology, Supervision, Project administration, Funding acquisition.
Lucia Passaro: Writing – review \& editing, Experiment Design, Data Analysis, Methodology.

\begin{appendices}

\section{Datasets} \label{ap:datasets}

In this section, we report the topic distribution of the document classification datasets for the administrative and legal domains, listed in Table \ref{tab:LabeldistrATTO} and Table \ref{tab:LabeldistrLegal}, respectively.

\begin{table*} 
\centering \small
\caption{Topic label distribution in the \textsc{atto} corpus.}
\label{tab:LabeldistrATTO}

\begin{tabular}{lc}
\toprule
\textbf{Topic} & \textbf{Num. of Docs} \\ 
\midrule
\textsc{Ambiente} (\textit{Environment}) & 296  \\ 
\textsc{Demografico} (\textit{Demographics})& 175  \\ 
\textsc{Avvocatura} (\textit{Advocacy}) & 117  \\ 
\textsc{Bandi e contratti} (\textit{Tenders and Contracts}) & 45  \\ 
\textsc{Commercio e attività economiche} (\textit{Trade and Economic activities})  & 8  \\ 
\textsc{Cultura, turismo e sport} (\textit{Culture, Tourism, and sport}) & 68  \\
\textsc{Edilizia} (\textit{Construction}) & 195 \\
\textsc{Personale} (\textit{Personnel}) & 209 \\
\textsc{Pubblica istruzione} (\textit{Education}) & 115 \\
\textsc{Servizi informativi} (\textit{Information services}) & 15 \\
\textsc{Servizio finanziario} (\textit{Financial services}) & 531 \\
\textsc{Sociale} (\textit{Welfare})  & 175 \\
\textsc{Urbanistica} (\textit{Urban planning})  & 863 \\ 
\midrule
\textit{Total} & \textit{2,812} \\
\bottomrule
\end{tabular}
\end{table*}

\begin{table*} 
\centering \small
\caption{Topic labels distribution in the legal document classification dataset.}
\label{tab:LabeldistrLegal}
\begin{tabularx}{1\textwidth}{lc}
\toprule
\textbf{Topic} & \textbf{Num. of Docs} \\ 
\midrule
\textsc{Polizia Locale} (\textit{Local Police})     & 688 \\
\textsc{Edilizia e Urbanistica} (\textit{Constructions and Urban Planning})               & 594 \\
\textsc{Attività economiche} (\textit{Economic Activities})                      & 405 \\
\textsc{Appalti e contratti} (\textit{Procurements and Contracts })               & 378 \\
\textsc{Servizi demografici} (\textit{Demographic Services})                     & 294 \\
\textsc{Tributi locali} (\textit{Local Taxes})                              & 255 \\
\textsc{Personale} (\textit{Personnel})                                & 224 \\
\textsc{Bilancio e contabilità} (\textit{Budget and Accounting})                    & 218 \\
\textsc{Amm. e segreteria generale} (\textit{Admin. and General Secretariat})   & 134 \\ 
\midrule
\textit{Total} & \textit{3,190} \\
\bottomrule
\end{tabularx}
\end{table*}

\section{Label words} \label{ap:labelwords}

This section shows the word labels encompassed within the different verbalizers we tested. Table \ref{et-verbalizer-legal} lists the words used in the base verbalizer for the legal entity typing experiment, while Table \ref{docclass-verbalizer-legal} reports those used for the document classification task. 

Table \ref{ap:knowverbaET-eng} and Table \ref{ap:knowverbaDC-eng} contain the English translation of some of the label words used in the models' knowledgeable verbalizers for the document classification and entity typing tasks in both PA and legal domains. They are the English translation of Table \ref{tab:knowverbaET} and Table \ref{tab:knowverbaDC}. 

Table \ref{docclass-verbalizer-en} contains the English version of the base and manual verbalizers adopted for the document classification in the administrative domain (see Table \ref{docclass-verbalizer} for the Italian version). 

\begin{table}
 \small
\caption{\label{et-verbalizer-legal} The label words used in the base verbalizer for the prompt entity typing experiments in the legal domain.}
\centering
\begin{tabularx}{\linewidth}{cl}

\toprule
 \textbf{Class} & \textbf{Base Verbalizer} \\

\midrule
\textsc{Jdg} & \textit{giudice} (\textit{judge}) \\
\midrule
 \textsc{Law} & \textit{legge} (\textit{law}) \\
\midrule
 \textsc{Lwy} & \textit{avvocato} (\textit{lawyer}) \\
\midrule 
 \textsc{Rul} & \textit{sentenza} (\textit{judgement}) \\
\bottomrule
\end{tabularx}
\end{table}

\begin{table*}[!ht]
\caption{\label{docclass-verbalizer-legal} The label words used in the base verbalizer for the prompt-based document classification experiments in the legal domain. }
\centering \small
\setlength{\tabcolsep}{5pt}
\renewcommand\tabularxcolumn[1]{m{#1}}
\begin{tabularx}{1\textwidth}{@{}ll@{}}
\textbf{Class} & \textbf{Base Verbalizer} \\
\midrule
\textsc{Polizia locale} & \textit{polizia, locale} (\textit{police, local}) \\
\midrule
\textsc{Edilizia e urbanistica} & \textit{edilizia, urbanistica} (\textit{constructions, urban planning}) \\
\midrule
\textsc{Attività economiche} & \textit{economia} (\textit{economy}) \\
\midrule
\textsc{Appalti e contratti} & \textit{appalti, contratti} (\textit{procurements, contracts}) \\
\midrule
\textsc{Servizi demografici} & \textit{demografia} (\textit{demographics}) \\
\midrule
\textsc{Tributi locali} & \textit{tributi} (\textit{taxes}) \\
\midrule
\textsc{Personale} & \textit{personale} (\textit{personnel}) \\
\midrule
\textsc{Bilancio e contabilità} & \textit{bilancio, contabilità} (\textit{budget, accounting}) \\
\midrule
\textsc{Amm. e segreteria generale} & \textit{amministrazione, segreteria} (\textit{administration, secretariat}) \\
\bottomrule
\end{tabularx}
\end{table*}

\begin{table}[h]
\centering
\caption{Some of the label words in the knowledgeable verbalizers of Ita-Legal-BERT, BureauBERTo, and UmBERTo constructed for the experiments on prompt entity typing. This is an English translation of Table \ref{tab:knowverbaET}. }
\label{ap:knowverbaET-eng}
\begin{tabular}{lllccc}
\toprule
\multirow{2}{*}{\textbf{Domain}} & \multirow{2}{*}{\textbf{Class}} & \multicolumn{3}{c}{\textbf{Knowledgeable Verbalizer}} \\
\cmidrule(lr){3-5}
& & Ita-Legal-BERT & BureauBERTo & UmBERTo \\
\midrule 
\multirow{2}{*}{\textbf{PA}} & \begin{tabular}[c]{@{}c@{}} \textsc{law} \end{tabular} & \begin{tabular}[c]{@{}c@{}} \textit{Regulation,} \\ \textit{directive,} \\ \textit{legislation,} \\ \textit{call for bids,} \\ \textit{norm,} \\ \textit{reform, ...}\end{tabular} & \begin{tabular}[c]{@{}c@{}} \textit{Regulation,} \\ \textit{Decree,} \\ \textit{Law,} \\ \textit{deliberation,} \\ \textit{norm,} \\ \textit{decrees, ...}\end{tabular} & \begin{tabular}[c]{@{}c@{}} \textit{contract,} \\ \textit{certificate,} \\ \textit{Protocol,} \\ \textit{code,} \\ \textit{legislation,} \\ \textit{title, ...}\end{tabular} \\
\\
& \textsc{OPA} & \begin{tabular}[c]{@{}c@{}} \textit{office,} \\ \textit{department,} \\ \textit{Responsible,} \\ \textit{coordinator,} \\ \textit{department,} \\ \textit{Institute, ...}\end{tabular} & \begin{tabular}[c]{@{}c@{}} \textit{service,} \\ \textit{sector,} \\ \textit{post,} \\ \textit{department,} \\ \textit{Department,} \\ \textit{institute, ...}\end{tabular} & \begin{tabular}[c]{@{}c@{}} \textit{field,} \\ \textit{service,} \\ \textit{operator,} \\ \textit{branch,} \\ \textit{department,} \\ \textit{sector, ...}\end{tabular} \\
\midrule

\multirow{2}{*}{\textbf{Legal}} & \begin{tabular}[c]{@{}c@{}} \textsc{LWY} \end{tabular} & \begin{tabular}[c]{@{}c@{}} \textit{defender,} \\ \textit{administrator,} \\ \textit{Av,} \\ \textit{male party,} \\ \textit{female party,} \\ \textit{assisted, ...}\end{tabular} & \begin{tabular}[c]{@{}c@{}} \textit{Avv,} \\ \textit{administrator,} \\ \textit{office,} \\ \textit{agent,} \\ \textit{defender,} \\ \textit{lawyers, ...}\end{tabular} & \begin{tabular}[c]{@{}c@{}} \textit{employee,} \\ \textit{operator,} \\ \textit{company,} \\ \textit{agency,} \\ \textit{legal,} \\ \textit{Avv, ...}\end{tabular} \\
\\
& \begin{tabular}[c]{@{}c@{}} \textsc{RUL} \end{tabular} & \begin{tabular}[c]{@{}c@{}} \textit{arrest,} \\ \textit{pronouncement,} \\ \textit{device,} \\ \textit{minutes,} \\ \textit{sentences,} \\ \textit{conviction, ...}\end{tabular} & \begin{tabular}[c]{@{}c@{}} \textit{determination,} \\ \textit{doctrine,} \\ \textit{judgment,} \\ \textit{Court,} \\ \textit{cause,} \\ \textit{Jurisprudence, ...}\end{tabular} & \begin{tabular}[c]{@{}c@{}} \textit{question,} \\ \textit{judgment,} \\ \textit{decisions,} \\ \textit{minutes,} \\ \textit{citation,} \\ \textit{order, ...}\end{tabular} \\

\bottomrule
\end{tabular}
\end{table}

\begin{table}[h]
\centering
\caption{Some of the label words in the knowledgeable verbalizers of Ita-Legal-BERT, BureauBERTo, and UmBERTo constructed for the experiments on prompt document classification. This is an English translation of Table \ref{tab:knowverbaDC}.}

\label{ap:knowverbaDC-eng}
\begin{tabular}{lllccc}
\toprule
\multirow{2}{*}{\textbf{Domain}} & \multirow{2}{*}{\textbf{Class}} & \multicolumn{3}{c}{\textbf{Knowledgeable Verbalizer}} \\
\cmidrule(lr){3-5}
& & Ita-Legal-BERT & BureauBERTo & UmBERTo \\
\midrule
\multirow{2}{*}{\textbf{PA}} & \begin{tabular}[c]{@{}c@{}} \textsc{Trade and} \\ \textsc{economic} \\ \textsc{activities}\end{tabular} & \begin{tabular}[c]{@{}c@{}} \textit{shares,} \\ \textit{performances,} \\ \textit{obligations,} \\ \textit{sales,} \\ \textit{taxes,} \\ \textit{transactions, ...}\end{tabular} & \begin{tabular}[c]{@{}c@{}} \textit{activities,} \\ \textit{professions,} \\ \textit{licenses,} \\ \textit{businesses,} \\ \textit{associations,} \\ \textit{shops, ...}\end{tabular} & \begin{tabular}[c]{@{}c@{}} \textit{activities,} \\ \textit{utilities,} \\ \textit{exercises,} \\ \textit{works,} \\ \textit{benefits,} \\ \textit{equipment, ...}\end{tabular} \\
\\
& \textsc{Advocacy} & \begin{tabular}[c]{@{}c@{}} \textit{advice,} \\ \textit{lawyer,} \\ \textit{judgments,} \\ \textit{court,} \\ \textit{defenders,} \\ \textit{judge,  ...}\end{tabular} & \begin{tabular}[c]{@{}c@{}} \textit{decree,} \\ \textit{procedure,} \\ \textit{decision,} \\ \textit{archiving,} \\ \textit{act,} \\ \textit{judgment, ...}\end{tabular} & \begin{tabular}[c]{@{}c@{}} \textit{decree,} \\ \textit{resolution,} \\ \textit{delegation,} \\ \textit{definition,} \\ \textit{delegated,} \\ \textit{deliberation, ...}\end{tabular} \\
\midrule

\multirow{2}{*}{\textbf{Legal}} & \begin{tabular}[c]{@{}c@{}} \textsc{Local police} \end{tabular} & \begin{tabular}[c]{@{}c@{}} \textit{surveillance,} \\ \textit{monitoring,} \\ \textit{traffic,} \\ \textit{road,} \\ \textit{guard,} \\ \textit{emergency, ...}\end{tabular} & \begin{tabular}[c]{@{}c@{}} \textit{verification,} \\ \textit{discipline,} \\ \textit{surveillance,} \\ \textit{Prefecture,} \\ \textit{Guard,} \\ \textit{security, ...}\end{tabular} & \begin{tabular}[c]{@{}c@{}} \textit{prevention,} \\ \textit{surveillance,} \\ \textit{verification,} \\ \textit{guard,} \\ \textit{monitoring,} \\ \textit{committee, ...}\end{tabular} \\
\\
& \begin{tabular}[c]{@{}c@{}} \textsc{Procurement } \\ \textsc{ and contracts} \end{tabular} & \begin{tabular}[c]{@{}c@{}} \textit{bid,} \\ \textit{contract,} \\ \textit{agreement,} \\ \textit{tender,} \\ \textit{price,} \\ \textit{sale, ...}\end{tabular} & \begin{tabular}[c]{@{}c@{}} \textit{form,} \\ \textit{contract,} \\ \textit{renewal,} \\ \textit{agreement,} \\ \textit{compromise,} \\ \textit{preliminary, ...}\end{tabular} & \begin{tabular}[c]{@{}c@{}} \textit{disciplinary,} \\ \textit{code,} \\ \textit{text,} \\ \textit{notice,} \\ \textit{treaty,} \\ \textit{decree, ...}\end{tabular} \\

\bottomrule
\end{tabular}
\end{table}

\begin{table*}[!ht]
\centering \small
\caption{\label{docclass-verbalizer-en} The Table shows the label adopted in experiments related to Document Classification in the PA domain.  This is an English translation of Table \ref{docclass-verbalizer}. Although some Italian words are translated as multi words word labels can be represented as single words only. 
}
\setlength{\tabcolsep}{5pt}
\renewcommand\tabularxcolumn[1]{m{#1}}
\begin{tabularx}{\textwidth}{@{}llX@{}}
\toprule
\makecell{\textbf{Class}} & \makecell{\textbf{Base Verbalizer}} & \textbf{Manual Verbalizer}\\
\midrule
\midrule
\makecell{\textsc{Environment}} & \makecell{environment} & \RaggedRight{environment, nature, land, flora, fauna, animals, climate, pollution, waste, hygiene, hunting, fishing, green, ecology, agriculture, water}\\
\midrule
\makecell{\textsc{Advocacy}} & \makecell{advocacy} & \RaggedRight{advocacy, attorneys, justice, legal, appeal, judges, courthouse, court, appello, assise, notification, acts, albo, pretorio, protocol}\\
\midrule
\makecell{\textsc{Tenders-Contracts}} & \makecell{tenders,\\contracts} & \RaggedRight{tenders, contracts, notice, contract, tender, hiring, liquidation}\\
\midrule
\makecell{\textsc{Trade-}\\\textsc{Economic-}\\\textsc{Activities}} & \makecell{trade,\\ economic,\\ activities} & \RaggedRight{trade, economy, business, economic, goods, trade, sales, purchases, merchants, confesercenti}\\
\midrule
\makecell{\textsc{Culture-}\\\textsc{Turism-}\\\textsc{Sport}} & \makecell{culture,\\ turism,\\ sport} & \RaggedRight{culture, tourism, sports, cultural, tourists, museums, art, cinema, vacations, entertainment, school, events}\\
\midrule
\makecell{\textsc{Demographic}} & \makecell{demographic} & \RaggedRight{demographics, population, inhabitants, residents, census, registry, residence, domicile, citizenship, conscription}\\
\midrule
\makecell{\textsc{Construction}} & \makecell{construction} & \RaggedRight{construction, building, yard, renovation, planimetry, residential}\\
\midrule
\makecell{\textsc{Personnel}} & \makecell{personnel} & \RaggedRight{personnel, resources, human, hiring, work, part-time}\\
\midrule
\makecell{\textsc{Education}} & \makecell{education} & \RaggedRight{education, institute, school, teacher, training, education}\\
\midrule
\makecell{\textsc{Information-}\\\textsc{Services}} & \makecell{services,\\ information} & \RaggedRight{services, information, informative}\\
\midrule
\makecell{\textsc{Financial-}\\\textsc{Services}} & \makecell{finance} & \RaggedRight{finance, euro, financial, accounting, accountant, coverage, refunds, payments, disbursements, budget, expenses, penalties, fines, taxes, wages, emoluments}\\
\midrule
\makecell{\textsc{Welfare}} & \makecell{welfare} & \RaggedRight{welfare, conscription, military, disabled, protection, civilian, disability}\\
\midrule
\makecell{\textsc{Urban-Planning}} & \makecell{urban planning} & \RaggedRight{urban planning, transportation, transports, traffic, circulation, vehicles, roadway}\\
\bottomrule
\end{tabularx}

\end{table*}

\section{Ablation study} \label{ap:ablation}

In this section, we report in Table \ref{tab:as-results} the results obtained in the ablation study on the entity typing task for the administrative domain. 

\begin{table*}
\centering \small
\caption{\label{tab:as-results} Ablation study conducted on BureauBERTo and UmBERTo on entity typing task by using the manual verbalizer. In bold are the best results for each entity class.}
\begin{tabular}{@{}lccccccccc@{}}
\toprule
\textbf{Model} &   & \textbf{\textsc{LOC}} & \textbf{\textsc{ORG}} & \textbf{\textsc{PER}} & \textbf{\textsc{ACT}} & \textbf{\textsc{LAW}} & \textbf{\textsc{OPA}} & \textbf{MicAvg} & \textbf{MacAvg} \\

 \midrule
\multirow{3}{*}{UmB}  & P & 0.79 & 0.29 & 0.48 & 0.63 & 0.73 & 0.54 & 0.57 & 0.58 \\
 & R & 0.37 & 0.27 & 0.58 & 0.28 & 0.81 & 0.54 & 0.48 & 0.47 \\
 & F1 & \textbf{0.50} & \textbf{0.28} & \textbf{0.53} & 0.39 & 0.76 & 0.54 & \textbf{0.52} & 0.50 \\
 \midrule

\multirow{3}{*}{BB}  & P & 0.75 & 0.28 & 0.39 & 0.63 & 0.79 & 0.75 & 0.57 & 0.60 \\
 & R & 0.30 & 0.22 & 0.45 & 0.42 & 0.78 & 0.57 & 0.46 & 0.46 \\
 & F1 & 0.43 & 0.25 & 0.42 & \textbf{0.50} & \textbf{0.79} & \textbf{0.65} & 0.51 & \textbf{0.51} \\

 \bottomrule
\end{tabular}
\end{table*}

\section{Fill-mask results} \label{ap:fillmask}

\begin{figure}[!htb]
  \centering \small
  \includegraphics[width=\linewidth]{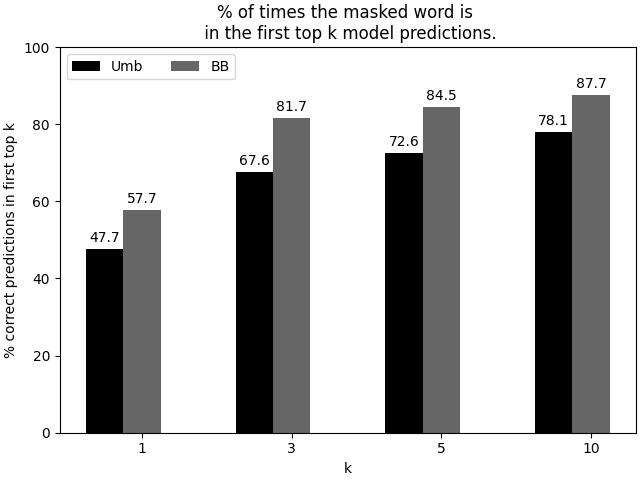}
  \caption{Results of a fill-mask task experiment in which Auriemma et al., (2023) masked domain-specific words in sentences from the ATTO corpus (PA domain). Percentages indicate the number of times the masked word was in the model's top k predictions.} 
  \label{fig:fillmask}
\end{figure}

Preliminary experiments on a fill-mask task (Fig.\ref{fig:fillmask}) showed that BureauBERTo outperformed UmBERTo when predicting masked words on Public Administration documents~\cite{auriemma2023bureauberto}. This motivated us to evaluate BureauBERTo domain-specific knowledge in an unsupervised setting in prompt-based zero-shot classification tasks.

\end{appendices}

\clearpage

\bibliography{paper-bibliography}

\end{document}